\documentclass[11pt]{article} 

\usepackage{subcaption} 
\usepackage{enumitem} 
\usepackage{mathtools}
\usepackage{amssymb} 
\usepackage{mathrsfs}

\usepackage{algorithm}
\usepackage{algpseudocode}
\usepackage{pifont}

\usepackage[para]{footmisc}

\usepackage{pgfplots}
\pgfplotsset{width=7.5cm,compat=1.9}
\usepackage{tikz}
\usetikzlibrary{shapes.geometric,arrows}
\usetikzlibrary{positioning,arrows.meta,shapes,automata,decorations.pathreplacing} 

\usepackage{todonotes}
\usepackage{hyperref}
\usepackage{xcolor}
\hypersetup{
	colorlinks,
	linkcolor={red},
	citecolor={green!90!black},
	urlcolor={blue!80!}
}
\usepackage{natbib}
 \bibpunct[, ]{(}{)}{,}{a}{}{,}%
\usepackage[all,arc]{xy}
\usepackage{mathrsfs}
\usepackage{latexsym,bm,amsmath,amsthm,graphicx,amssymb,dsfont}
\usepackage[utf8]{inputenc}
\usepackage[english]{babel}
\usepackage{mathrsfs}
\usepackage{mathtools}
\usepackage{graphicx}

\usepackage{graphicx}
\setlist[enumerate]{itemsep=0mm}
\usepackage{tikz}
\usepackage{algorithm}
\usepackage{pifont}

\usepackage{natbib}

\usepackage{todonotes}

\usepackage{hyperref} 

\newtheorem{thm}{Theorem}[section]
\newtheorem{corollary}[thm]{Corollary}
\newtheorem{proposition}[thm]{Proposition}
\newtheorem{lemma}[thm]{Lemma}

\theoremstyle{definition}
\newtheorem{theorem}{Theorem}
\newtheorem*{definition}{Definition}

\theoremstyle{definition}

\newtheorem{example}[thm]{Example}

\newtheorem*{remark}{Remark}

\theoremstyle{remark}

\newtheorem{assumption}{Assumption}
\usepackage{etoolbox}
\makeatletter
\patchcmd{\@maketitle}
{\ifx\@empty\@dedicatory}
{\ifx\@empty\@date \else {\vskip3ex \centering\footnotesize\@date\par\vskip1ex}\fi
	\ifx\@empty\@dedicatory}
{}{}
\patchcmd{\@adminfootnotes}
{\ifx\@empty\@date\else \@footnotetext{\@setdate}\fi}
{}{}{}
\makeatother

\usepackage[margin=1in]{geometry}

\usepackage{fancyhdr} 

\makeatletter
\let\runauthor\@author
\let\runtitle\@title

\usepackage[toc,page]{appendix}

\title{Thompson Sampling for Infinite-Horizon Discounted Decision Processes}
\usepackage[T1]{fontenc}
\usepackage[utf8]{inputenc}
\usepackage{authblk}

\author[1]{Daniel Adelman}
\author[2]{Cagla Keceli}
\author[3]{Alba V. Olivares Nadal}

\affil[1]{Booth School of Business, The University of Chicago, Chicago, IL 60637, USA}
\affil[1]{United Airlines Chicago, IL 60605, USA}
\affil[3]{UNSW Business School, UNSW, Kensington NSW 2033, Australia}

\begin{document}
\date{}
\maketitle

\begin{abstract}
This paper develops a framework for learning in discounted infinite-horizon Markov decision processes (MDPs) with Borel state and action spaces, whose rewards and transitions depend on an unknown parameter.
To analyze sampling-based adaptive learning algorithms in this setting, we introduce a canonical probability space that explicitly incorporates sampled parameters into the history of the process. 
As a performance criterion, we adopt the per-period suboptimality gap used in discounted-MDP regret analysis and specialize it to our parametrized Bayesian setting. Since this quantity captures the remaining loss in future performance from the current period onward, we refer to it as \emph{residual regret}. We use the expected residual regret to connect discounted-MDP regret analysis with asymptotic discount optimality from adaptive control and the temporal-difference error perspective from reinforcement learning. We then focus on Thompson sampling (TS) in discounted infinite-horizon MDPs. Under assumptions that extend those used in prior work on finite state and action spaces to the Borel setting, we show that the expected residual regret for TS converges to zero exponentially fast. We further show that, under mild conditions ensuring the existence of the relevant limits, the \textcolor{black}{sample-path} residual regret converges to zero almost surely and TS achieves complete learning.

\end{abstract}




%


\section{Introduction}

We consider a discounted infinite-horizon Markov decision process (MDP) in which a decision-maker (DM) operates under uncertainty about a parameter $\theta$ governing both reward and transition dynamics. We denote these parametrized MDPs, which need not be Markovian, as $\theta$-MDPs. At each period, given the current state and chosen action, the system generates a random reward and transitions to a random next state, with distributions determined by the unknown parameter. The DM uses observed outcomes to adapt decisions sequentially. In seeking to maximize cumulative reward without knowledge of the underlying parameter governing rewards and state transitions, the DM faces a fundamental exploration-exploitation trade-off: whether to select actions that provide information about uncertain parameter values or actions that are expected to yield high rewards. This trade-off is commonly addressed by adaptive, sampling-based learning algorithms, which maintain and update beliefs about the unknown parameter and select actions accordingly.


In this paper, we develop a general canonical probability space for sampling-based adaptive learning algorithms and \textcolor{black}{discuss regret metrics suitable for discounted infinite-horizon settings}. 
We then turn to Thompson sampling (TS) \cite{thompson1933likelihood}, a classical benchmark among sampling-based algorithms. TS was first analyzed in the context of multi-armed bandit  problems, a degenerate class of MDPs. The extensive literature on TS for bandits establishes both finite-time regret guarantees and asymptotic convergence properties. For example, \cite{agrawal2012analysis} shows that the expected regret of TS grows logarithmically with the time horizon. 

More recently, TS has become one of the standard Bayesian approaches for
reinforcement learning, where the environment is typically modelled as an MDP. In this context, TS is
also referred to as posterior sampling for reinforcement learning. At each resampling time, TS draws
an MDP from the current posterior distribution and implements a policy that is
optimal for the sampled model. The foundational analyses of TS focus on
finite-horizon or episodic MDPs\footnote{For episodic finite-horizon MDPs, the time horizon can be chosen to coincide with the episode length \cite{osband2013more,ouyang2017learning}} and establish regret guarantees for finite state and action spaces
\cite{osband2013more,osband2017why}. TS remains an active
benchmark in finite-horizon MDPs, with recent work extending its
analysis to continuous state and control problems by imposing additional structure:
\cite{li2024prior} considers linear-mixture MDPs,
\cite{moradipari2023improved} uses a finite discretization of the environments space, and \cite{fan2021model,bayrooti2026noregret} consider state and action
spaces that are subsets of finite-dimensional real spaces, together with
Gaussian-process or linear-Gaussian models and additional regularity
conditions.

 
A smaller body of work studies TS in infinite-horizon MDPs. 
Most of these analyses assume finite state and action spaces and adopt a long-run average-cost criterion
\cite{kim2017thompson,gopalan2015thompson,ouyang2017learning,agrawal2017optimistic,gimelfarb2023uninformative}. Their regret guarantees 
typically rely on structural conditions on the controlled process: for example,
\cite{kim2017thompson} assumes an ergodicity condition,
\cite{gopalan2015thompson} imposes recurrence conditions on a designated
starting state, \cite{ouyang2017learning} considers weakly communicating MDPs and dynamic episodes, \cite{agrawal2017optimistic} considers communicating MDPs with finite
diameter, and \cite{gimelfarb2023uninformative} assumes that the optimal-policy chain is ergodic unichain. Only a few infinite-horizon TS analyses permit more general,
potentially infinite state or action spaces. Among the few exceptions, \cite{banjevic2019thompson} allows general measurable state, action, and parameter spaces, while \cite{abeille2018improved} considers state and action spaces that are subsets of $\mathbb{R}^{d_s}$ and $\mathbb{R}^{d_a}$, respectively, where $\mathbb{R}$ denotes the set of real numbers. The latter further assumes a linear-quadratic control model. Both consider the average-cost criterion. To the best of our knowledge, no existing work establishes regret guarantees for TS in discounted infinite-horizon MDPs with Borel state and action spaces. Our results close this specific theoretical gap. Similarly to
\cite{kim2017thompson,gimelfarb2023uninformative}, our regret guarantees rely
on informativeness conditions based on the Kullback-Leibler divergence. Our assumption imposes uniform positivity of the transition density, which rules out nontrivial absorbing regions and implies accessibility across the state space.



\subsection{Contributions}

\paragraph {\bf We develop a canonical probability space for sampling-based adaptive learning algorithms that explicitly incorporates the sequence of sampled parameters into the history of the process} Although \cite{kim2017thompson} and \cite{banjevic2019thompson} formulate TS on a similar probabilistic framework, the sampled parameters are not included as coordinates of the sample space. To the best of our knowledge, no previous work explicitly augments the history with the sequence of sampled parameters.
This unified probabilistic framework facilitates the analysis of sampling-based algorithms by placing sampled parameters, controls, rewards, and states on the same probability space.
In particular, the formulation makes evident that the underlying adaptive learning process is inherently history-dependent and therefore not Markovian. To emphasize this distinction, we refer to the resulting process with an unknown parameter as a $\theta$-MDP rather than simply an MDP.

\textbf{We connect the per-period suboptimality gap underlying the discounted-MDP regret in \cite{zhou} with asymptotic discount optimality (ADO) from adaptive control \cite{manfred1987estimation,hernandez2012adaptive} and the Bellman-error perspective underlying temporal-difference (TD) learning \cite{sutton2018reinforcement,hernandez2012adaptive}.}
In particular, we refer to this suboptimality gap as \emph{residual regret}. Because it is a random expectation, obtained by conditioning on the random history (which includes the sampled parameter), we also refer to this term as {\it sample-path residual regret} or {\it probabilistic residual regret}. We show that, in the present parametrized Bayesian setting, a policy is ADO if and only if its expected residual regret vanishes. We further show that the expected residual regret admits a representation in terms of the TD error function. To the best of our knowledge, these connections have not previously been exploited to analyze TS in parametrized discounted MDPs with Borel state and action spaces.

\paragraph {\bf We demonstrate that the expected residual regret of TS converges to 0 exponentially fast}  Our work mostly complements \cite{kim2017thompson} and \cite{banjevic2019thompson}. Both works show good performance guarantees for TS under parameter uncertainty, yet our work has a stronger connection with the former since we also assume a finite parameter space. The latter adopts a continuous parameter space, similar to \cite{hernandez2012adaptive}, which results in a setting that is harder to analyze. We extend the result of \cite{kim2017thompson} to a more general setting with Borel state and action spaces and a discounted infinite-horizon criterion, by adapting their assumptions to this broader framework.

\paragraph {\bf We show that the posterior belief on the true parameter converges to one almost surely (a.s.) under TS} 
This property, commonly referred to as \emph{complete learning}, is stronger than convergence in expectation, as it provides guarantees along individual sample paths rather than only on average. While convergence of posterior beliefs has been studied extensively in bandit settings, most existing results focus on expected posterior behavior. Complete learning has been analyzed in bandit contexts \cite{freedman1963asymptotic}, and average-reward MDPs with finite state and action spaces \cite{kim2017thompson}. To our knowledge, it has not been studied in MDPs with Borel state and action spaces or under discounted reward criteria. In the adaptive learning literature, alternative methods such as the minimum contrast estimator are known to achieve complete learning \cite{hernandez2012adaptive}, but these results do not apply to sampling-based algorithms such as TS. By connecting complete learning to the concept of ADO, we provide a probabilistic counterpart to this notion.

\paragraph {\bf We prove that the probabilistic residual regret of TS converges to 0 a.s}  This provides a sample-path guarantee that strengthens our results on expected residual regret. While the latter establishes convergence in expectation, this result ensures that residual regret vanishes along almost every realization of the learning process.


\subsection{Structure of the paper}
The paper is organized as follows. We begin with the adaptive learning framework and performance criteria: Section~\ref{modelSetup} constructs the canonical probability space for sampling-based policies and Section~\ref{sec:regret} defines residual regret and connects its expected version to ADO and TD errors. In Section \ref{ThompsonSampling}, we characterize TS by defining the posterior update and control selection mechanisms. In Section \ref{regretBound}, we provide an asymptotic analysis of the expected residual regret. In Section \ref{completeLearning}, we show that TS exhibits complete learning, i.e., the posterior sampling error converges to 0 a.s. We also show that sample-path residual regret converges to 0 a.s.



\section{Model Setup\label{modelSetup}} 

In this section, we introduce the parametrized MDP framework and establish the notation used throughout the paper. Section \ref{MDP} formulates the MDP when the DM knows $\theta$, following \cite{hernandez2012adaptive}. From Section \ref{ProblemFormulation} onward, $\theta$ is assumed unknown, and the DM solves the $\theta$-MDP using estimates. We first outline a general sampling algorithm and then construct the associated canonical probability space over histories, accounting for the sampled parameters $\theta_t$. 

Table \ref{tab:notation_model_setup} summarizes the main notation introduced in this section. Throughout the paper, we reserve upper-case letters to denote random variables, lower-case letters to realizations of random variables, script letters to spaces and sometimes sets, and bold upper-case letters to elements of $\sigma$-algebras to be defined. 


\begin{table}[htbp]
\begin{small}
\centering
\renewcommand{\arraystretch}{0.75}
\begin{tabular}{p{0.15\textwidth} p{0.85\textwidth}}
\hline
\textbf{Notation} & \textbf{Meaning}\\
\hline
$\mathbb{R}$ & The set of real numbers\\
$\mathcal{B}(\cdot)$ & Borel $\sigma$-algebra on the corresponding topological space. \\
$\mathcal{X}$ & Borel state space. \\

$\mathcal{U}$ & Borel control (action) space. \\

$\mathcal{U}(x)$ & Set of admissible controls at state $x\in\mathcal{X}$. \\

$\mathbb{K}$ & Set of admissible state-control pairs (Eq. \eqref{eq:K})\\

$\mathcal{P}$ & Finite parameter space\\
$\mathscr{R}_c$ & Compact measurable reward space, with $\mathscr{R}_c\subset \mathbb{R}_+$. \\

$\beta$ & Discount factor, with $\beta\in[0,1)$. \\

$\theta$ & True (unknown) parameter in $\mathcal{P}$ governing rewards and transitions \\

$\gamma$ & Generic parameter value in $\mathcal{P}$. \\

$X_t$ & Random state in period $t$. \\

$U_t$ & Random control applied in period $t$. \\

$R_t$ & Random reward generated in period $t$. \\

$\Theta_t$ & Random parameter sample drawn in period $t$ by the sampling algorithm. \\

$x_t,u_t,r_t,\theta_t$ & Realizations of $X_t,U_t,R_t,\Theta_t$, respectively. \\

$\lambda$ & $\sigma$-finite reference measure on $(\mathscr{R}_c,\mathcal{B}(\mathscr{R}_c))$ (Eq. \eqref{exp_reward}). \\


$r^\theta(\cdot,\cdot)$ & Expected one-step reward under parameter $\theta$ (Eq. \eqref{exp_reward}). \\
$f^\theta(\cdot\mid x,u)$ & One-step reward density under parameter $\theta$ given $(x,u)$ (Eq. \eqref{exp_reward}). \\

$\eta$ & $\sigma$-finite reference measure on $(\mathcal{X},\mathcal{B}(\mathcal{X}))$ (Eq. \eqref{eq:Q}). \\
$Q^\theta(\cdot\mid x,u)$ & State-transition distribution under parameter $\theta$ given $(x,u)$ (Eq. \eqref{eq:Q}). \\
$q^\theta(\cdot\mid x,u)$ & One-step transition density under parameter $\theta$ given $(x,u)$ (Eq. \eqref{eq:Q})\\

$\nu^\theta(\cdot)$ & Optimal value function for the true parameter $\theta$ (Eqs. \eqref{optimalityEqn} and \eqref{v(x)}). \\

$\mu^\theta$ & Optimal policy associated with parameter $\theta$.\\ 

$\overline{\mathcal{H}}_t$ & 
Space of histories up to period $t$ (Eq. \eqref{eq:ovHt}). \\

$\mathcal{H}_t$ & Space of {\it admissible} histories up to period $t$, subspace of  $\overline{\mathcal{H}}_t$ (Eq. \eqref{eq:Ht}). \\

$H_t$ & Random admissible history up to period $t$ (Eq. \eqref{history_vector}).\\

$h_t$ & Realization of the history random variable $H_t$. \\

$\Omega=\overline{\mathcal{H}}_\infty$ & Sample space of infinite-horizon histories (Eq. \eqref{eq:Omega}).  \\
$\omega \in \Omega$ & Sample path (Eq. \eqref{eq:omega}).\\


$\pi_t(\cdot\mid H_t)$ & Posterior distribution on $\mathcal{P}$ at period $t$, conditional on history $H_t$; $\theta_t$ is sampled from this distribution (Def. \ref{def:pit}).\\

$\mu=\{\mu_t\}$ & Randomized history-dependent control policy, where each $\mu_t$ is a stochastic kernel on $\mathcal{U}$ given $\mathcal{H}_t$ and $\mathcal{P}$ (Def. \ref{def:mu}). \\

$\mathcal{M}$ & Set of admissible randomized policies (i.e., $\mu_t \in \mathcal{M}$ for all $t$). \\

$\mathbb{P}_{x_0}^{\mu,\theta}$ & Prob. measure induced by policy $\mu$, initial state $x_0$, and true parameter $\theta$. \\

$\mathbb{E}_{x_0}^{\mu,\theta}$ & Expectation operator with respect to $\mathbb{P}_{x_0}^{\mu,\theta}$. \\

$E_{f^\theta}[\cdot\mid x,u]$ & One-step expectation with respect to the reward density $f^\theta(\cdot\mid x,u)$. \\

$E_{q^\theta}[\cdot\mid x,u]$ & One-step expectation with respect to the transition density $q^\theta(\cdot\mid x,u)$. \\

$E_{f^\theta q^\theta}[\cdot\mid x,u]$ & One-step expectation with respect to both reward and next-state randomness under parameter $\theta$. \\


$V_{x_0}^{\mu,\theta}(n)$ & Infinite-horizon expected discounted reward from period $n$ onward under policy $\mu$, when the process starts from $x_0$ at period $0$ (Eq. \eqref{V(n)}). \\

\hline
\end{tabular}
\caption{Notation introduced in the model setup. \label{tab:notation_model_setup}}
\end{small}
\end{table}

\subsection{Markov Decision Process for Known $\theta$ \label{MDP}} 


We consider a discrete-time MDP $(\mathcal{X},\mathcal{U},\{\mathcal{U}(x),x\in\mathcal{X}\},f^\theta,q^\theta)$, where the reward and transition densities depend on a parameter $\theta\in\mathcal{P}$, with $\mathcal{P}$ a finite parameter space. The components of this MDP are as follows:

\begin{enumerate}[leftmargin=*]
\item \textit{State space} $\mathcal{X}$, a Borel space. The system random state at time $t\in\{0,1,2,\dots\}$ is denoted by $X_t\in\mathcal{X}$, and its realization by $x_t$.

\item \textit{Control space} $\mathcal{U}$, a Borel space. The random control applied at time $t$ is $U_t\in\mathcal{U}$, whereas its realization is denoted by $u_t$.

\item \textit{Set of admissible controls} $\mathcal{U}(x)\subseteq \mathcal{U}$, a compact set for each $x\in\mathcal{X}$. We assume $\mathcal{U}=\bigcup_{x\in\mathcal{X}}\mathcal{U}(x)$ and denote the set of admissible state-control pairs as
\begin{equation}\label{eq:K}
\mathbb{K}\coloneqq \{(x,u)\mid x\in\mathcal{X},\,u\in\mathcal{U}(x)\}.
\end{equation}
We assume $\mathbb{K}$ is measurable in $\mathcal{X}\times\mathcal{U}$.

\item \textit{Reward.} Given $(x,u)\in\mathbb{K}$, the system generates a random reward $R_t\in\mathscr{R}_c\subset\mathbb{R}_+$, where $\mathscr{R}_c$ is compact and measurable according to the conditional distribution of rewards $F^\theta(\cdot\mid x,u)$. We denote the observed realization of this reward as $r_t$. The distribution of rewards admits a measurable, continuous, one-step reward density\footnote{Radon-Nikodym derivative of $F^\theta$ with respect to $\lambda$.} with respect to a $\sigma$-finite measure $\lambda$ on $(\mathscr{R}_c,\mathcal{B}(\mathscr{R}_c))$, so that
\[
F^\theta(\mathbf{R}\mid x,u)=\int_{\mathbf{R}} f^\theta(r\mid x,u)\,d\lambda(r), \quad \forall \mathbf{R}\in\mathcal{B}(\mathscr{R}_c).
\]
The expected reward is
\begin{align} \label{exp_reward}
r^\theta(x,u)=\int_{\mathscr{R}_c} r\,f^\theta(r\mid x,u)\,d\lambda(r).
\end{align}

\item \textit{State transition.} Given a state-action pair $(x,u)\in\mathbb{K}$ in period $t$, the next state $X_{t+1}$ follows the distribution $Q^\theta(\cdot\mid x,u)$, which admits a one-step transition density\footnote{Radon-Nikodym derivative of $Q^\theta$ with respect to $\eta$} $q^\theta(\cdot\mid x,u)$ with respect to a $\sigma$-finite measure $\eta$ on $(\mathcal{X},\mathcal{B}(\mathcal{X}))$, so that:
\begin{equation} \label{eq:Q}
Q^\theta(\mathbf{X}\mid x,u)=\int_{\mathbf{X}} q^\theta(y\mid x,u)\,d\eta(y), \quad \forall \mathbf{X}\in\mathcal{B}(\mathcal{X}).
\end{equation}
Equivalently,
\[
Q^\theta(\mathbf{X}\mid x,u)=\mathbb{P}(X_{t+1}\in \mathbf{X}\mid x_t=x,u_t=u), \quad \forall \mathbf{X}\in\mathcal{B}(\mathcal{X}).
\]
\end{enumerate}

\begin{remark} \label{boundedExpReward}
	Since $f^\theta(\cdot\mid x_t,u_t)$ is continuous on a compact set $\mathscr{R}_c$, it attains its minimum and maximum in $\mathscr{R}_c$. Thus, there exists a constant $M\ge 0$ such that $\vert r^\theta(x_t,u_t)\vert\le M$ $\forall (x_t,u_t)\in\mathbb{K}$. 
	To see why this holds, note that 
	\begin{align*}
	\vert r^\theta(x_t,u_t)\vert \le \int_{\mathscr{R}_c} \vert r f^\theta(r\mid x_t,u_t)\vert d\lambda(r) 
	&\le \int_{\mathscr{R}_c} \max (\vert r f^\theta(r\mid x_t,u_t)\vert) d\lambda(r)
	\\
	&= \max_{r\in\mathscr{R}_c} \vert r f^\theta(r\mid x_t,u_t)\vert \lambda(\mathscr{R}_c) < \infty,
	\end{align*}
	where the first inequality follows by (\ref{exp_reward}) and Jensen's inequality, and the equality holds since $\max (\vert r f^\theta(r\mid x_t,u_t)\vert)$ is constant in $r$. Then, $\max\limits_{r\in\mathscr{R}_c} \vert r f^\theta(r\mid x_t,u_t)\vert$ is finite because $f^\theta(\cdot\mid x_t,u_t)$ is continuous on a compact set, and $\lambda(\mathscr{R}_c)$ is finite since $\mathscr{R}_c\subset \mathbb{R}_{+}$ is compact and $\lambda$ is a $\sigma$-finite measure.
\end{remark}

The DM aims to maximize the infinite-horizon expected total discounted reward. The \emph{optimal value function} $\nu^\theta(x)$ represents the maximum such reward starting from state $x$, and it depends on the true parameter $\theta$. It is known from \cite{hernandez2012discrete} that, under some assumptions, this problem is solved by the Bellman equation 
\begin{equation} \label{optimalityEqn}
\nu^\theta(x) \coloneqq \sup_{u\in\mathcal{U}(x)}\left\{r^\theta(x,u) + \beta\int_{y\in\mathcal{X}} \nu^\theta(y) dQ^\theta(dy\mid x,u) \right\}, \quad \forall x\in\mathcal{X}, \theta\in\mathcal{P},
\end{equation}
where $\beta\in [0,1)$ is the discounting factor. If $\nu^\theta(x)$ is a solution to (\ref{optimalityEqn}), let $\mu^\theta$ denote the corresponding optimal policy for parameter $\theta$. Under sufficient technical conditions, $\mu^\theta$ is stationary, deterministic, and Markovian. As verifying these conditions is outside the scope of this paper, we make the following assumption.

\begin{assumption} \label{uniqueSoln} 
	For all $\theta\in\mathcal{P}$, there exists a unique solution to (\ref{optimalityEqn}) such that the supremum is attained, and there exists an optimal policy $\mu^\theta$ that is stationary, deterministic, and Markovian.
\end{assumption}

\begin{remark}
	Assumption \ref{uniqueSoln} is necessary for all of the main results of the paper. Accordingly, we do not explicitly refer to it in subsequent lemma, proposition, or theorem statements.
\end{remark}


\subsection{Adaptive Learning with Sampling \label{ProblemFormulation}} 

Had $\theta$ been known, the DM would maximize the objective function by solving (\ref{optimalityEqn}) under Assumption \ref{uniqueSoln}. From this point onward, we treat $\theta$ as an unknown but fixed parameter. A control problem with an unknown parameter is referred to as an \emph{adaptive} control problem; see \cite{hernandez2012adaptive} for a formal definition.

When $\theta$ is unknown, the problem becomes one of learning, in which the DM must choose controls while gathering information about $\theta$. We focus on \emph{sampling-based algorithms}, in which the DM maintains a belief distribution $\pi_t$, samples a parameter $\Theta_t$, and selects a control $U_t$ based on the information available up to time $t$, referred to as the \emph{history}. To formalize this framework, we first introduce the notions of history, beliefs, and policies. We then describe the sampling-based learning procedure, construct the associated canonical probability space, and define the performance criterion used to evaluate policies.

\paragraph{\bf Histories, Beliefs and Policies} We denote the space of histories at any period of time as $\overline{\mathcal{H}}_t$. For $t = 0$, we have $\overline{\mathcal{H}}_0\coloneqq \mathcal{X}$. For $t\ge 1$, the space of histories has product form,

\begin{equation}\label{eq:ovHt}
\overline{\mathcal{H}}_t\coloneqq (\mathcal{X}\times\mathcal{P}\times\mathcal{U}\times\mathscr{R}_c)^t\times\mathcal{X} = \overline{\mathcal{H}}_{t-1} \times \mathcal{P}\times \mathcal{U}\times\mathscr{R}_c\times \mathcal{X}.
\end{equation}

The history random variable $H_t\in \overline{\mathcal{H}}_t$ at period $t\in\mathbb{N}_{\ge2}$ follows the recursion
\begin{align} \label{history_vector}
H_t\coloneqq (X_0,\Theta_0,U_0,R_0,\dots,X_{t-1},\Theta_{t-1},U_{t-1},R_{t-1},X_t),
\end{align}
The history vector's realization is denoted by
$$
h_t\coloneqq (x_0,\theta_0,u_0,r_0,\dots,x_{t-1},\theta_{t-1},u_{t-1},r_{t-1},x_t),
$$
where $x_t,\ \theta_t, \ u_t$, are the realizations of $X_t$, $U_t$ and $\Theta_t$, respectively.

We denote the space of {\it admissible} histories up to period $t$ by $\mathcal{H}_t$. For $t=0$ we have that $\mathcal{H}_0 \coloneqq   \overline{\mathcal{H}}_0= \mathcal{X}$, and for $ t\ge 1$, $\mathcal{H}_t$ is a subspace of  $\overline{\mathcal{H}}_t$. In particular

\begin{equation}\label{eq:Ht}
\mathcal{H}_t\coloneqq \left \{ H_t \in \overline{\mathcal{H}}_t: (X_t,U_t)\in \mathbb{K} \right \}
\end{equation}
The history spaces $\mathcal{H}_0$ and $\mathcal{H}_t$ ($t=1,2,\dots$) are endowed with their product Borel $\sigma$-algebras $\mathcal{B}(\mathcal{X})$, $\mathcal{B}(\mathcal{P})$, $\mathcal{B}(\mathcal{U}(x))$ and $\mathcal{B}(\mathscr{R}_c)$.

We are now equipped to formally introduce the concepts of belief distributions and randomized policies in our sample-based learning setting.
\begin{definition}\label{def:pit}
	The distribution $\pi_t(\cdot\mid H_t): \mathcal{P}\mid \mathcal{H}_t\to \mathbb{R}_{+}$ is a belief distribution over $\Theta_t\in\mathcal{P}$. It is a function of the random and time-dependent history $H_t$. 
\end{definition}

\begin{definition}\label{def:mu}
	A randomized policy $\mu = \{\mu_t\}$ is a sequence of stochastic kernels $\mu_t$ on $\mathcal{U}$ given $\mathcal{H}_t$ and $\mathcal{P}$, satisfying

\begin{equation*}
\mu_t(\mathcal{U}(x_t)\mid h_t,\theta_t)= 1 \text{ for all } h_t\in\mathcal{H}_t, \, \theta_t\in\mathcal{P}, \text{ and } t\ge 0.
\end{equation*}
The stochastic kernel $\mu_t$ is an element of the set of admissible control policies, denoted by $\mathcal{M}$.
\end{definition}


\paragraph{\bf Overview of a Sampling Algorithm\label{overview}}  

\begin{figure}[htb!]
	\begin{center}
		\begin{tikzpicture}[%
		every node/.style={
			font=\normalsize, 
			text height=1ex,
			text depth=.25ex,
		},
		]
		\draw[thick,->] (0,0) -- (13,0);
		\foreach \x in {2.5,4.7,5,7.5,10,12.5}{
			\draw (\x cm,3pt) -- (\x cm,0pt);
		}
		\node[align=center] at (4.7,1.85) {$\pi_t(\cdot\mid H_t)$};
		\draw [->] (4.7,1.4) -- (4.7,0.15);
		\node[align=center] at (10,1.85) {$f^\theta(\cdot\mid X_t,U_t)$};
		\draw [->] (10,1.4) -- (10,0.15);
		\node[align=center] at (12.5,1.85) {$q^\theta(\cdot\mid X_t,U_t)$};
		\draw [->] (12.5,1.4) -- (12.5,0.15);
		\draw (0,0) node[below=3pt] {} node[above=3pt] {};
		\draw (2.5,0) node[below=3pt] {$X_t$} node[above=3pt] {};
		\draw (5,0) node[below=3pt] {$\theta_t$} node[above=3pt] {};
		\draw (7.5,0) node[below=3pt] {$U_t$} node[above=3pt] {};
		\draw (10,0) node[below=3pt] {$R_t$} node[above=3pt] {};
		\draw (12.5,0) node[below=3pt] {$X_{t+1}$} node[above=3pt] {};
		\draw [thick,decorate,decoration={brace,amplitude=6pt,raise=0pt}] 
		(0.1,0.1) -- (2.4,0.1);
		\node[align=center] at (1.25,0.5) {$H_t$};
		\draw [thick,decorate,decoration={brace,amplitude=6pt,raise=0pt,mirror}] 
		(0.1,-0.6) -- (12.4,-0.6);
		\node[align=center] at (6.25,-1.25) {$H_{t+1}$};
		\end{tikzpicture}
		\caption{Evolution of the stochastic process, in the case when $\theta$ is not known.} \label{stochastic_timeline}
	\end{center}
\end{figure}
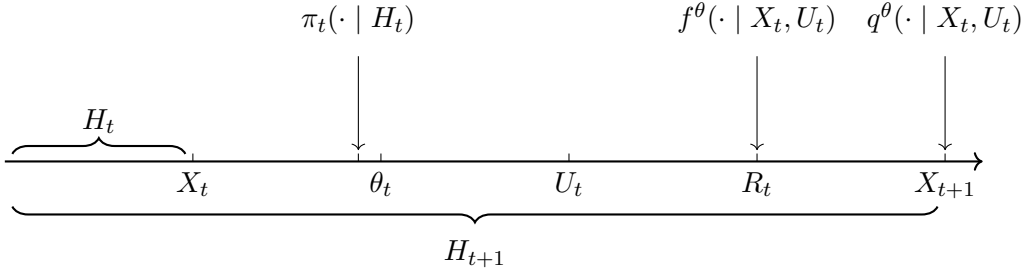

Figure \ref{stochastic_timeline} illustrates the random variables driving the learning process in their order of occurrence,
and Algorithm \ref{alg:sampling} provides a pseudocode for a general sampling algorithm.

\begin{algorithm}[H]
\caption{Adaptive Learning via Sampling \label{alg:sampling} }
\begin{algorithmic}[1]
\State Observe initial state $X_0$. Initialize $H_0=X_0$ and belief distribution $\pi_0(\cdot \mid H_0)$
\For{$t = 0,1,2,\dots$}
    \State Observe history $H_t$ and state $X_t$
    \State Compute belief $\pi_t(\cdot \mid H_t)$
    \State Sample $\Theta_t \sim \pi_t(\cdot \mid H_t)$
    \State Select action $U_t \sim \mu_t(\cdot \mid H_t, \Theta_t)$
    \State Observe $R_t \sim f^\theta(\cdot \mid X_t,U_t)$
    \State Observe $X_{t+1} \sim q^\theta(\cdot \mid X_t,U_t)$
    \State Update history $H_{t+1}$
    \State Update belief $\pi_{t+1}(\cdot \mid H_{t+1})$
\EndFor
\end{algorithmic}
\end{algorithm}

At the beginning of period $t$, the DM observes the history $H_t$, which contains all information up to and including the current state $X_t$. The function $\pi_t(\cdot\mid H_t)$, provided by the system designer, maps $H_t$ to a distribution over the sampled parameter $\Theta_t$. Although $\pi_t(\cdot\mid H_t)$ is deterministic given $H_t$, it is random due to the randomness of $H_t$. The DM then draws $\Theta_t$ according to $\pi_t(\cdot\mid H_t)$. The randomness of $\Theta_t$ arises both from the randomness of $H_t$ and from the sampling procedure itself. Given the realized sample $\theta_t$, the DM selects a control $U_t\in\mathcal{U}(X_t)$ by solving the corresponding optimization problem. 
The state-control pair $(X_t, U_t)$ generates a reward $R_t \sim f^\theta(\cdot \mid X_t, U_t)$ and a next state $X_{t+1} \sim q^\theta(\cdot \mid X_t, U_t)$, both governed by the true parameter $\theta$. The augmented tuple $(\Theta_t, U_t, R_t, X_{t+1})$ extends the history from $H_t$ to $H_{t+1}$, after which the DM updates the belief to $\pi_{t+1}(\cdot \mid H_{t+1})$. Conditional on $(X_t, U_t)$, the reward and next state are independent.\footnote{Yet, there can be alternative settings where an exogenous random variable impacts both $R_t$ and $X_{t+1}$, resulting in dependence between the reward and the next state.}


Although the true parameter $\theta$ governs the observed process (i.e. rewards and transitions are generated according to $f^\theta(\cdot\mid x,u)$ and $q^\theta(\cdot\mid x,u)$), for some sampling algorithms, the DM will need access to the reward and transition densities associated with a parameter $\gamma \in \mathcal{P}$, $f^\gamma(\cdot\mid x,u)$ and $q^\gamma(\cdot\mid x,u)$. Henceforth, we assume that $f^\gamma$ and $q^\gamma$ are available for any $\gamma\in\mathcal{P}$.

\paragraph{\bf Canonical Construction\label{formulation}}

The infinite sequence of four-tuples $\overline{\mathcal{H}}_\infty$ is the sample space, denoted by $\Omega$,
\begin{equation}\label{eq:Omega}
\Omega = \overline{\mathcal{H}}_\infty \coloneqq
(\mathcal{X}\times\mathcal{P}\times\mathcal{U}\times\mathscr{R}_c)^\infty = \mathcal{X}\times\mathcal{P}\times\mathcal{U}\times\mathscr{R}_c\times\mathcal{X}\times\mathcal{P}\times\mathcal{U}\times\mathscr{R}_c\dots
\end{equation}

$\Omega$ is the space of histories $\overline{H}_t = (X_0,\Theta_0,U_0,R_0,X_1,\Theta_1,U_1,R_1,\dots)$ with $X_t\in\mathcal{X}$, $\Theta_t\in\mathcal{P}$, $U_t\in\mathcal{U}$, $R_t\in\mathscr{R}_c$ for $t\ge0$. The state, parameter, control and reward variables are defined as projections from $\Omega$ to sets $\mathcal{X}$, $\mathcal{P}$, $\mathcal{U}$ and $\mathscr{R}_c$, respectively. 

A typical element of the sample space, $\omega\in\Omega$, is an infinite sequence of the form below:
\begin{equation}\label{eq:omega}
\omega = (x_0,\theta_0,u_0,r_0,x_1,\theta_1,u_1,r_1,\dots), \  x_t\in \mathcal{X} \text{, } \theta_t\in\mathcal{P} \text{, } u_t \in\mathcal{U} \text{, } r_t\in\mathscr{R}_c, \ \forall \, t\ge 0.
\end{equation}

We use $\mathcal{B}(\Omega)$ to denote the corresponding product $\sigma$-algebra of $\Omega$. The final element of the probability space is the probability measure $\mathbb{P}_{x_0}^{\mu,\theta}:\mathcal{B}(\Omega)\to[0,1]$. It represents the probability measure when policy $\mu\in\mathcal{M}$ is used, the initial state is $X_0 = x_0$, and the true parameter is $\theta$. The expectation operator with respect to $\mathbb{P}_{x_0}^{\mu,\theta}$ is $\mathbb{E}_{x_0}^{\mu,\theta}$. Whenever this expectation is taken, we take the random variables $(X_n,\Theta_n,U_n,R_n)_{\forall n}$ as generated by $\mathbb{P}_{x_0}^{\mu,\theta}$. We have a collection of sample paths that are parametrized by $\mu$, $\theta$ and $x_0$. The operands of these operators are specified by the underlying state process induced by $\mu$, $\theta$ and $x_0$. We emphasize that the space of admissible histories, $\mathcal{H}_t$, is contained in $\Omega = \overline{\mathcal{H}}_\infty$, and therefore, the (admissible) history random variable $H_t$ is defined on $(\Omega,\mathcal{B}(\Omega),\mathbb{P}_{x_0}^{\mu,\theta})$.


A randomized, history-dependent policy $\mu$ induces a probability measure $\mathbb{P}_{x_0}^{\mu,\theta}$ on $(\Omega,\mathcal{B}(\Omega))$. By the Ionescu-Tulcea Theorem, proved in Proposition 7.28 of \cite{shreve1978stochastic}, for any given policy $\mu=\{\mu_t\}\in\mathcal{M}$, any initial state $X_0=x_0$ and true parameter $\theta\in\mathcal{P}$, there exists a unique probability measure $\mathbb{P}_{x_0}^{\mu,\theta}$ on $(\Omega,\mathcal{B}(\Omega))$, satisfying: 

\begin{enumerate}[label=(\alph*),leftmargin=*]
	\item $\mathbb{P}_{x_0}^{\mu,\theta}(\mathcal{H}_\infty) = 1$,
	\item $\mathbb{P}_{x_0}^{\mu,\theta}(X_0=x_0) = 1$,
	\item $\mathbb{P}_{x_0}^{\mu,\theta}(\Theta_t = \theta_t\mid h_t) = \pi_t(\theta_t\mid h_t)$ for all $\theta_t\in\mathcal{P}$ \textnormal{given} $h_t\in\mathcal{H}_t$ and $t\ge 0$,
	\item $\mathbb{P}_{x_0}^{\mu,\theta}(U_t\in \mathbf{U}\mid h_t,\theta_t) = \mu_t(\mathbf{U}\mid h_t,\theta_t)$ for all $\mathbf{U}\in\mathcal{B}(\mathcal{U})$ \textnormal{given} $h_t\in\mathcal{H}_t$, $\theta_t\in\mathcal{P}$, and $t\ge 0$,
	\item $\mathbb{P}_{x_0}^{\mu,\theta}(R_t\in\mathbf{R}\mid h_t,\theta_t,u_t)= \int_\mathbf{R} f^\theta(r\mid x_t,u_t) d\lambda(r)$ for all $\mathbf{R}\in\mathcal{B}(\mathscr{R}_c)$ \textnormal{given} $h_t\in\mathcal{H}_t$, $\theta_t\in\mathcal{P}$, $u_t\in\mathcal{U}(x_t)$, and $t\ge 0$. When conditioned on $x_t$ and $u_t$, $R_t$ is independent of $\theta_t$ and $h_t \setminus \{x_t\}$.
	\item $\mathbb{P}_{x_0}^{\mu,\theta}(X_{t+1}\in \mathbf{X}\mid h_t,\theta_t,u_t,r_t) = \int_\mathbf{X} q^\theta(y\mid x_t,u_t) d\eta(y)$ for all $\mathbf{X}\in \mathcal{B}(\mathcal{X})$ \textnormal{given} $h_t\in\mathcal{H}_t$, $\theta_t\in\mathcal{P}$, $u_t\in\mathcal{U}(x_t)$, $r_t\in\mathscr{R}_c$, and $t\ge 0$. When conditioned on $x_t$ and $u_t$, $X_{t+1}$ is independent of $\theta_t$, $r_t$ and $h_t \setminus \{x_t\}$.
\end{enumerate}

The probability measure $\mathbb{P}_{x_0}^{\mu,\theta}$ induced by the policy $\mu$ satisfies all (a)-(f), where 

\begin{enumerate}[label=(\alph*), itemsep = 0.01mm]
	\item is by the definition of probability measure. 
	\item is by construction, it implies the initial state of the process is $x_0$ with probability 1. 
	\item shows the history-dependent posterior distribution, from which the sample $\theta_t$ is generated.
	\item is the decision rule, i.e., the collection of policies. In each period $t$, the DM selects controls not only by the current state $x_t$, but by the entire history vector $h_t$.
	The decision rule also depends on the sample $\theta_t$. Since $h_t$ ends with $x_t$, we add $\theta_t$ as a condition. 
	\item characterizes the random reward drawn from the distribution which knows the noisy version of $\theta$. Given $x_t$ and $u_t$, the random reward $R_t$, generated from density $f^\theta(\cdot\mid X_t,U_t)$, does not depend on the sample $\theta_t$. Since $\theta_t$ does not give any additional information, it can be dropped. If not conditioned on $x_t$ and $u_t$, then $R_t$ depends on $h_t$, $\theta_t$, and $u_t$. 
	\item is the state transition law. Set $\mathbf{X}$ represents the states that are accessible from $x_t$. Given $x_t$ and $u_t$, the random next state $X_{t+1}$, generated from density $q^\theta(\cdot\mid x_t,u_t)$, is independent of $\theta_t$ and $r_t$. Since $\theta_t$ and $r_t$ are superfluous, they can be dropped. If not conditioned on $x_t$ and $u_t$, then $X_{t+1}$ depends on $h_t$, $\theta_t$, $u_t$ and $r_t$.
\end{enumerate}

When the expectation is over one step instead of the entire process, we use a different notations, i.e., different than $\mathbb{E}_{x_0}^{\mu,\theta}$. If the expectation is taken with respect to the random reward, we denote it by $E_{f^\theta}[\cdot\mid x,u]$, where density $f^\theta$ is with respect to $\sigma$-finite measure $\lambda$. If the expectation is with respect to the random next state, the operator is $E_{q^\theta}[\cdot\mid x,u]$, where density $q^\theta$ is with respect to the $\sigma$-finite measure $\eta$. If it is with respect to both the random reward and the random next state, then we use $E_{f^\theta q^\theta}[\cdot\mid x,u]$. Whenever there is a $\cdot$ in these expectation operators, the $\cdot$ implies a random variable. 

\paragraph{\bf Objective Function (Performance Criteria)\label{adaptivePolicy}} 
Given the initial state $x_0$ and the discount factor $\beta\in(0,1)$, the expected discounted reward over the infinite-horizon of implementing a policy $\mu$ from period $t=0$ onward is 
\begin{equation} \label{V(0)}
V_{x_0}^{\mu,\theta}(0) \coloneqq \mathbb{E}_{x_0}^{\mu,\theta} \left[\sum_{t=0}^\infty\beta^{t}R_t\right], \quad \forall \mu\in \mathcal{M}, \, x_0\in \mathcal{X}.
\end{equation}

When the rewards from period $0$ up until $n-1$ are dismissed and the discounting starts from period $n$ onward, we have
\begin{equation} \label{V(n)}
V_{x_0}^{\mu,\theta}(n) \coloneqq \mathbb{E}_{x_0}^{\mu,\theta} \left[\sum_{t=n}^\infty\beta^{t-n}R_t\right].
\end{equation}

As in (\ref{V(0)}), the DM implements policy $\mu$, starting from (known) state $x_0$ in period 0. However, the rewards in (\ref{V(n)}) are accumulated only from period $t=n$ onward, when the (random) system state is $X_n$. We will use the notation $n$ when we fix a specific time period and $t$ when we take summations.

For a given policy, the DM should know the value of $\theta$ to compute (\ref{V(0)}) and (\ref{V(n)}). Alternatively, the DM can statistically estimate these quantities if they have access to the reward density and transition density of the oracle to simulate the policy. 

Recall from (\ref{optimalityEqn}) that the optimal value function of the standard MDP problem is $\nu^\theta(x)$, with optimal policy $\mu^\theta$. In the adaptive setting, when the DM is assumed to know $\theta$, the optimal policy is $\mu^\theta$;  we call it the $\theta$-optimal policy (which, under Assumption \ref{uniqueSoln}, is stationary, deterministic, and Markovian). Hence, without loss of generality, we define $\nu^\theta(x_0)$ with respect to $\mathbb{E}_{x_0}^{\mu,\theta}$, i.e.,
\begin{equation} \label{v(x)}
\nu^{\theta}(x_0) \coloneqq \sup_{\mu\in \mathcal{M}} V_{x_0}^{\mu,\theta}(0) = \sup_{\mu\in\mathcal{M}}\mathbb{E}_{x_0}^{\mu,\theta} \left[\sum_{t=0}^\infty \beta^tR_t\right] = \mathbb{E}_{x_0}^{\mu^\theta,\theta} \left[\sum_{t=0}^\infty \beta^tR_t\right], \quad \forall x_0\in \mathcal{X},
\end{equation}
where $\nu^{\theta}(x_0)$ is equal to $\nu^\theta(x)$ in (\ref{optimalityEqn}) when $x=x_0$, $\forall x\in\mathcal{X}$.



 \section{Metrics of Regret\label{sec:regret}}


To evaluate the performance of an admissible policy, we adopt the notion of per-period suboptimality employed by \cite{zhou} in their study of reinforcement learning for discounted MDPs. There, regret is defined as the cumulative sum of these suboptimality gaps over time. We adapt this notion to our parametrized Bayesian setting and refer to it as residual regret.

\begin{definition} \label{def:residual_regret}
The residual regret
	$\mathbb{R}_{x_0}^{\mu,\theta}(n)$ is a random expectation that represents the forward-looking regret from period $n$ onward into the infinite future. This regret is between a policy $\mu$ until it switches to the optimal policy $\mu^\theta$ in period $n$ as opposed to continuing with $\mu$. Formally,
	\begin{align}
	\mathbb{R}_{x_0}^{\mu,\theta}(n) &\coloneqq
	\nu^\theta(X_n) - \mathbb{E}^{\mu,\theta}_{x_0}\left[\sum_{t=n}^\infty\beta^{t-n}R_t \mid H_n \right] \label{resRegDiff1} \\
	&= \mathbb{E}_{x_0}^{\mu^\theta,\theta}\left[ \sum_{t=n}^\infty \beta^{t-n} R_t \mid H_n \right] - \mathbb{E}_{x_0}^{\mu,\theta}\left[\sum_{t=n}^\infty\beta^{t-n}R_t \mid H_n\right], \label{resRegDiff3}
	\end{align}
	where the conditioning on the random history vector is denoted explicitly. 
\end{definition}

In what follows, we use the terms {\it probabilistic residual regret}, {\it sample-path residual regret} and {\it residual regret} interchangeably. The residual regret is a random expectation, and is obtained by conditioning on the random history $H_n$, generated by running a sampling-based policy $\mu$ from period 0 to $n$.  The term $\mathbb{R}_{x_0}^{\mu,\theta}(0)$ corresponds to the per-period suboptimality gap studied by \cite{zhou} in the present parametrized Bayesian setting. We next define its expectation.

\begin{definition} \label{def:expected_residual_regret} 
The expected residual regret, i.e., $\mathcal{R}_{x_0}^{\mu,\theta}(n)$, is the expected forward-looking regret from period $n$ onward into the infinite future. This regret is between a policy which implements a policy $\mu$ until it switches to the optimal policy $\mu^\theta$ in period $n$ as opposed to continuing with $\mu$. Formally,
	\begin{equation} \label{ResidualRegret}
	\mathcal{R}_{x_0}^{\mu,\theta}(n) \coloneqq 
	\mathbb{E}_{x_0}^{\mu,\theta} [\nu^\theta(X_n)] - V_{x_0}^{\mu,\theta}(n).
	\end{equation} 
\end{definition}

By the law of iterated expectations, the expected residual regret is the expectation
of the residual regret, namely,
\begin{align}  \label{expectedAndProbabilisticRR}
\mathcal{R}_{x_0}^{\mu,\theta}(n)= \mathbb{E}_{x_0}^{\mu,\theta}[\mathbb{R}_{x_0}^{\mu,\theta}(n)].
\end{align}

Throughout the remainder of the paper, we use the expected residual regret $\mathcal{R}_{x_0}^{\mu,\theta}$ and its
sample-path counterpart $\mathbb{R}_{x_0}^{\mu,\theta}$ to evaluate the performance of TS.
For completeness, Appendix~\ref{stdRegret} relates these quantities to the
conventional discounted value gap by decomposing the expectation of the latter into three components:
the finite-time regret accumulated before period $n$, a state-induced regret that
captures the effect of reaching a suboptimal state, and the expected residual regret.
This decomposition shows that the expected residual regret is the only component
that remains under the DM's control from period $n$ onward.

In the remainder of this section, we relate the expected residual regret to two
notions from adjacent literatures. First, we show that vanishing expected residual
regret is equivalent, in our setting, to ADO a criterion
introduced in the adaptive control literature for discounted problems
\cite{manfred1987estimation}. Second, we express the expected residual regret in
terms of a TD error, connecting our performance criterion to the
Bellman-error perspective underlying temporal-difference methods for value-function
learning in MDPs \cite{sutton2018reinforcement}.

\subsection{Connection to ADO} \label{VRR}

\cite{hernandez2012adaptive} describes the idea behind ADO as ``to allow the system to run during a learning period of $n$ stages'' and defines an ADO policy as follows: 

\begin{definition}[ADO] 
	A policy $\mu$ is called \emph{asymptotically discount optimal} (ADO) if, 
	\begin{align*} 
	\left\vert V_{x_0}^{\mu,\theta}(n) - \mathbb{E}_{x_0}^{\mu,\theta} [\nu^\theta(X_n)]\right\vert \to 0 \textnormal{  as  } n\to \infty, \quad \forall x_0\in \mathcal{X},
	\end{align*}
	where $\nu^\theta(X_n)$ is defined by (\ref{v(x)}) and $V_{x_0}^{\mu,\theta}(n)$ by (\ref{V(n)}). 
\end{definition}

We label an ADO policy as $\theta$-ADO to emphasize the dependence on the underlying parameter $\theta$. Traditionally, expected regret is formulated without the absolute value \cite{lai1985asymptotically}. It follows that the expected residual regret is equivalent to the $\theta$-ADO expression, i.e.,
\begin{align} \label{noAbsValue}
\mathcal{R}^{\mu,\theta}_{x_0}(n)\coloneqq \mathbb{E}_{x_0}^{\mu,\theta}[\nu^\theta(X_n)] - V_{x_0}^{\mu,\theta}(n) = \vert V_{x_0}^{\mu,\theta}(n) - \mathbb{E}_{x_0}^{\mu,\theta}[\nu^\theta(X_n)]\vert.
\end{align}
Showing (\ref{noAbsValue}) will be instrumental in bounding the expected residual regret. 

\begin{lemma} \label{absoluteValue}
	The absolute value in the $\theta$-ADO expression can be omitted in our setting, i.e., 
	$$
	\vert V_{x_0}^{\mu,\theta}(n) - \mathbb{E}_{x_0}^{\mu,\theta}[\nu^\theta(X_n)]\vert = \mathbb{E}_{x_0}^{\mu,\theta}[\nu^\theta(X_n)] - V_{x_0}^{\mu,\theta}(n). 
	$$  
\end{lemma}

The statement in Lemma \ref{absoluteValue} is self-evidently true, but because the notation is cumbersome in this setting, we include the proof here.

\proof[Proof of Lemma \ref{absoluteValue}]
Consider the $\theta$-ADO statement, $\vert V_{x_0}^{\mu,\theta}(n) - \mathbb{E}_{x_0}^{\mu,\theta}[\nu^\theta(X_n)]\vert$. It can be rewritten as
\begin{align}
&\vert V_{x_0}^{\mu,\theta}(n) - \mathbb{E}_{x_0}^{\mu,\theta}[\nu^\theta(X_n)]\vert \\
&= \left\vert \mathbb{E}_{x_0}^{\mu,\theta} \sum_{t=n}^\infty\beta^{t-n}R_t - \mathbb{E}_{x_0}^{\mu,\theta}[\nu^\theta(X_n)]\right\vert \nonumber \\ 
&= \left \vert \mathbb{E}_{x_0}^{\mu,\theta} \left [\mathbb{E}_{x_0}^{\mu,\theta} \left(\sum_{t=n}^\infty\beta^{t-n}R_t \mid X_n\right) - \mathbb{E}_{x_0}^{\mu,\theta}[\nu^\theta(X_n)\mid X_n]\right]\right\vert \nonumber \\
&= \left\vert \mathbb{E}_{x_0}^{\mu,\theta} \left [\mathbb{E}_{x_0}^{\mu,\theta} \left(\sum_{t=n}^\infty\beta^{t-n}R_t \mid X_n\right) - \nu^\theta(X_n)\right]\right\vert \nonumber \\
&= \left\vert\mathbb{E}_{x_0}^{\mu,\theta} \left[\mathbb{E}_{x_0}^{\mu,\theta} \left(\sum_{t=n}^\infty\beta^{t-n}R_t(X_t,U_t) \mid X_n\right) - \sup_{\mu\in\mathcal{M}}\mathbb{E}_{X_n}^{\mu,\theta}\left[\sum_{t=0}^\infty\beta^t R_t(X_t,U_t)\right]\right]\right\vert. \label{negativeAbsolute}
\end{align} 

The second equality is by the law of iterated expectations, and the fourth is by substituting in the definition of the optimal value function. 
Evidently, the first term of (\ref{negativeAbsolute}) is upper bounded by the second term. Thus, (\ref{negativeAbsolute}) is equal to
\begin{align*}
\mathbb{E}_{x_0}^{\mu,\theta} \left[\sup_{\mu\in\mathcal{M}}\mathbb{E}_{X_n}^{\mu,\theta}\left[\sum_{t=0}^\infty\beta^t R_t(X_t,U_t)\right] - \mathbb{E}_{x_0}^{\mu,\theta} \left(\sum_{t=n}^\infty\beta^{t-n}R_t(X_t,U_t) \mid X_n\right)\right],
\end{align*}
for any initial state $x_0\in \mathcal{X}$. Rewriting the optimal value function in closed form, we obtain
\begin{align*}
\mathbb{E}_{x_0}^{\mu,\theta} \left[\nu^\theta(X_n) - \mathbb{E}_{x_0}^{\mu,\theta} \left(\sum_{t=n}^\infty\beta^{t-n}R_t(X_t,U_t) \mid X_n\right)\right],
\end{align*}
which is equal to
$$
\mathbb{E}_{x_0}^{\mu,\theta}[\nu^\theta(X_n)] - \mathbb{E}_{x_0}^{\mu,\theta} \left[\sum_{t=n}^\infty\beta^{t-n}R_t(X_t,U_t) \right]
= \mathbb{E}_{x_0}^{\mu,\theta}[\nu^\theta(X_n)] - V_{x_0}^{\mu,\theta}(n).
$$

\endproof

Rather than calling policies $\theta$-ADO, we say that they have ``vanishing expected residual regret''. 

\begin{definition} A policy $\mu$ has vanishing expected residual regret if, 
	$$
	\lim_{n\to\infty} \mathcal{R}^{\mu,\theta}_{x_0}(n)\coloneqq \lim_{n\to\infty} \left[\mathbb{E}_{x_0}^{\mu,\theta}[\nu^\theta(X_n)] - V_{x_0}^{\mu,\theta}(n)\right] = 0, \quad \forall x_0\in \mathcal{X},
	$$
	where $\theta$ is the true parameter and $X_n$ is the random period-$n$ state, which is obtained by running an admissible policy $\mu$ for $n$ periods.
\end{definition}

If sampling-based learning algorithm has vanishing expected residual regret, then eventually its expected performance converges to that of an optimal policy.

\subsection{Connection to TD Error}
The TD error function is parametrized by $\theta$ and quantifies the discrepancy between the reward-to-go of choosing the optimal control instead of an arbitrary control in a given state.  Following the Bellman-residual characterization used in the adaptive control literature \cite{hernandez2012adaptive}, we consider the definition below:

\begin{definition}[Hern{\'a}ndez-Lerma, 2012] \label{phiDefn}
	We denote the temporal-difference error function 
	by $\phi^\theta: \mathbb{K}\to \mathbb{R}$, where
	$$
	\phi^\theta(x,u) \coloneqq r^\theta(x,u) + \beta\int \nu^\theta(y) Q^\theta(dy \mid x,u) - \nu^\theta(x).
	$$ 
	The first two terms of $\phi^\theta(x,u)$ constitute the reward-to-go of choosing (an arbitrary) control $u\in\mathcal{U}(x)$ in state $x$, while $\nu^\theta(x)$ is the reward-to-go of choosing the optimal control in $x$.
\end{definition}


The following lemma provides a characterization of vanishing expected residual regret in terms of the TD error, thereby linking our regret criterion to the Bellman-error perspective used in reinforcement learning. Specifically, a policy has vanishing expected residual regret if and only if its TD error converges to zero in probability.

\begin{lemma}[Temporal-difference error] \label{phi}
	For every initial state $x_0\in\mathcal{X}$, a policy $\mu$ has vanishing expected residual regret, i.e.,
	$$\lim_{n\to\infty} \left[\mathbb{E}_{x_0}^{\mu,\theta}[\nu^\theta(X_n)] - V_{x_0}^{\mu,\theta}(n)\right] = 0, \quad \forall x_0\in \mathcal{X},$$
	if and only if $\phi^\theta(X_t,U_t)\to 0$ in probability-$\mathbb{P}_{x_0}^{\mu,\theta}$ for every $x_0\in \mathcal{X}$.
\end{lemma}

The proof of Lemma \ref{phi} is an adaptation of Theorem 3.6 in \cite{hernandez2012adaptive} to our setting. It establishes a connection between the expected residual regret and the expected value of $\phi^\theta$. In particular, for a policy $\mu$,
\begin{equation} \label{phi_residualregret}
\mathcal{R}^{\mu,\theta}_{x_0}(n) = -\sum_{t=n}^\infty \beta^{t-n} \mathbb{E}_{x_0}^{\mu,\theta} \phi^\theta(X_t,U_t). 
\end{equation}

\proof[Proof of Lemma \ref{phi}]
By the definition in Section \ref{VRR}, 
$$
\phi^\theta(X_t,U_t) = \mathbb{E}_{x_0}^{\mu,\theta}[R_t(X_t,U_t) + \beta \nu^\theta(X_{t+1}) - \nu^\theta(X_t)\mid H_t,U_t],
$$
for any initial state $x_0\in\mathcal{X}$, admissible policy $\mu\in\mathcal{M}$, true parameter value $\theta$, $t\ge 0$. For any $t\ge 1$, history in period $t$ is $h_t = (x_0,\theta_0,a_0,r_0,x_1,\theta_1,a_1,r_1,\dots,x_t)$. Multiplying by $\beta^{t-n}$ yields
$$
\beta^{t-n}\phi^\theta(X_t,U_t) = \mathbb{E}_{x_0}^{\mu,\theta}[\beta^{t-n}R_t(X_t,U_t) + \beta^{t-n+1}\nu^\theta(X_{t+1}) - \beta^{t-n}\nu^\theta(X_t)\mid H_t,U_t].
$$

Taking the expectation of both sides and by the law of total expectation, we obtain
$$
\mathbb{E}_{x_0}^{\mu,\theta} [\beta^{t-n}\phi^\theta(X_t,U_t)] = \mathbb{E}_{x_0}^{\mu,\theta} [\beta^{t-n}R_t(X_t,U_t) + \beta^{t-n+1}\nu^\theta(X_{t+1}) - \beta^{t-n}\nu^\theta(X_t)].
$$

Summing over all $t\ge n$ gives
\begin{align*}
\sum_{t=n}^\infty \beta^{t-n} \mathbb{E}_{x_0}^{\mu,\theta} [\phi^\theta(X_t,U_t)] &=  \sum_{t=n}^\infty \mathbb{E}_{x_0}^{\mu} [\beta^{t-n} R_t(X_t,U_t)] \\
&\qquad + \mathbb{E}_{x_0}^{\mu,\theta} \left[\sum_{t=n}^\infty (\beta^{t-n+1}\nu^\theta(X_{t+1}) - \beta^{t-n}\nu^\theta(X_t))\right].
\end{align*}

The above simplifies into 
\begin{align*} 
\sum_{t=n}^\infty \beta^{t-n} \mathbb{E}_{x_0}^{\mu,\theta} [\phi^\theta(X_t,U_t)] = V_{x_0}^{\mu,\theta}(n) - \mathbb{E}_{x_0}^{\mu,\theta}[\nu^\theta(X_n)].
\end{align*}

Recalling the result of Lemma \ref{absoluteValue} as,
\begin{align*} 
-\sum_{t=n}^\infty \beta^{t-n} \mathbb{E}_{x_0}^{\mu,\theta} [\phi^\theta(X_t,U_t)] =  \mathbb{E}_{x_0}^{\mu,\theta}[\nu^\theta(X_n)] - V_{x_0}^{\mu,\theta}(n).
\end{align*}

In the limit, $\mu$ has vanishing expected residual regret (is $\theta$-ADO); that is, for every $x_0\in\mathcal{X}$,
$$
\lim_{n\to\infty} (\mathbb{E}_{x_0}^{\mu,\theta}[\nu^\theta(X_n)] - V_{x_0}^{\mu,\theta}(n))= 0,
$$
if and only if, for every $x_0\in \mathcal{X}$, 
$$
\lim_{n\to\infty} -\sum_{t=n}^\infty \beta^{t-n} \mathbb{E}_{x_0}^{\mu,\theta} [\phi^\theta(X_t,U_t)] = 0.
$$

This is equivalent to, for every $x_0\in \mathcal{X}$,
$$
\lim_{t\to\infty} -\mathbb{E}_{x_0}^{\mu,\theta} [\phi^\theta(X_t,U_t)] = 0.
$$

Hence, a policy $\mu$ has vanishing expected residual regret if and only if, for every $x_0\in\mathcal{X}$,
$$
\lim_{t\to\infty} \mathbb{E}_{x_0}^{\mu,\theta} [\phi^\theta(X_t,U_t)] = 0.
$$

By Theorem 4.1.4 of \cite{chung2001course}, if $\mathbb{E}_{x_0}^{\mu,\theta} [\phi^\theta(X_t,U_t)]$ converges to 0, then $\phi^\theta(X_t,U_t)$ converges to 0 in probability-$\mathbb{P}_{x_0}^{\mu,\theta}$, for every $x_0\in \mathcal{X}$, proving the forward direction of Lemma \ref{phi}.
It remains to show the reverse direction to complete the proof. By the same theorem of \cite{chung2001course}, for a uniformly bounded sequence $\{\phi^\theta\}$, convergence in probability and $\mathcal{L}^p$ (in expectation) are equivalent. By the first remark in Section \ref{MDP}, 
it follows that $\{\phi^\theta\}$ is bounded by some finite number. Hence, $\phi^\theta(X_t,U_t)\to 0$ in probability-$\mathbb{P}_{x_0}^{\mu,\theta}$ implies $\mathbb{E}_{x_0}^{\mu,\theta} [\phi^\theta(X_t,U_t)]\to 0$ as $t\to0$.
\endproof



\section{Thompson Sampling\label{ThompsonSampling}}


The framework developed in the previous sections accommodates a broad class of sampling-based algorithms, characterized by a belief update rule and a control selection policy. Up to this point, we have considered a general sampling-based policy, denoted by $\mu=\{\mu_t\}$, together with a belief distribution $\pi_t(\cdot \mid H_t)$. From this point onward, we specialize to TS, which is the focus of the remainder of the paper. Under TS, the DM samples a parameter from the posterior distribution and selects a control that is optimal for the sampled parameter. We use the terms Thompson sampling, TS, TS algorithm, TS policy, and TS decision rule interchangeably. To distinguish TS from a general sampling-based policy $\mu$, we denote the TS policy by $\tau$. The probability measure induced by TS and its corresponding expectation operator are denoted by $\mathbb{P}_{x_0}^{\tau,\theta}$ and $\mathbb{E}_{x_0}^{\tau,\theta}$, respectively.

We next specify TS by defining its posterior update rule in Section \ref{PosteriorDistribution} and its control selection mechanism in Section \ref{DecisionRule}.


\subsection{Posterior Update\label{PosteriorDistribution}} 

The TS algorithm generates an estimate $\theta_t$ in each period, by using the ``synthetic'' belief update function $\pi_t(\theta_t\mid H_t)$. Initially, the DM holds the prior belief $\pi_0(\theta_0\mid h_0) > 0$ on the true parameter $\theta$. That is, the unknown $\theta$ is modeled by the $\vert \mathcal{P}\vert$-valued random variable $\Theta_t$, with initial prior distribution
$$
\pi_0(\theta_0\mid h_0) \coloneqq \mathbb{P}_{x_0}^{\tau,\theta}(\Theta_0 = \theta_0\mid h_0), \quad \forall \theta_0\in \mathcal{P}.
$$

At the beginning of period $t$, the DM updates her belief over the parameter candidates, by computing the (random) posterior distribution
\begin{align} \label{piisprobability}
\pi_t(\theta_t\mid H_t) \coloneqq \mathbb{P}_{x_0}^{\tau,\theta}(\Theta_t = \theta_t\mid H_t), \quad \forall \theta_t\in\mathcal{P}.
\end{align}

The expected value of $\pi_t(\theta_t\mid H_t)$,  
\begin{align*} 
\pi_t(\theta_t) \coloneqq \mathbb{E}_{x_0}^{\tau,\theta}[\pi_t(\theta_t\mid H_t)],
\end{align*}
is deterministic, and its dependence on $(x_0,\tau,\theta)$ is implicit. We employ Bayes' Theorem to conduct the update, 
\begin{equation} \label{BayesTheorem}
\pi_t(\theta_t\mid H_t) = \frac{\mathcal{L}_t^{\theta_t}(H_t) \pi_0(\theta_0\mid h_0)}{\sum\limits_{\gamma\in\mathcal{P}}\mathcal{L}_t^\gamma(H_t) \pi_0(\gamma\mid h_0)},
\end{equation}
where $\mathcal{L}_t^\gamma(H_t):\mathcal{H}_t\to\mathbb{R}_{ }$ is the (history-dependent) likelihood function. For any $\gamma\in\mathcal{P}$, 
\begin{align*} 
\mathcal{L}_t^\gamma(H_t) \coloneqq \prod_{s=1}^t f^\gamma(R_{s-1}\mid X_{s-1},U_{s-1})q^\gamma(X_s\mid X_{s-1},U_{s-1}).
\end{align*}

The joint density $f^\gamma(\cdot\mid x,u)q^\gamma(\cdot\mid x,u)$ specifies a joint probability measure on $[0,1]\times \mathcal{X}$, 
$$
\rho^{\gamma}_{x,u}(\mathbf{R},\mathbf{X})\coloneqq \int_\mathbf{R} f^\gamma(r\mid x,u) d\lambda(r) \int_\mathbf{X} q^\gamma(y\mid x,u) d\eta(y),
$$
for $\mathbf{R}\subseteq\mathcal{B}(\mathscr{R}_{c}),\,\mathbf{X}\subseteq \mathcal{B}(\mathcal{X})$. 
Then, for any parameter value $\gamma\in\mathcal{P}$, the ratio of the Radon-Nikodym derivative is
$$
\frac{d\rho^\theta_{x,u}}{d\rho^\gamma_{x,u}} = \frac{f^\theta(\cdot\mid x,u)q^\theta(\cdot\mid x,u)}{f^\gamma(\cdot\mid x,u)q^\gamma(\cdot\mid x,u)}.
$$

\noindent We use this ratio to introduce the concept of relative entropy.

\begin{definition}
	The relative entropy of $\rho^\theta_{x,u}$ with respect to $\rho^\gamma_{x,u}$ is
	$$
	\mathcal{K}(\rho^\theta_{x,u}\mid \rho^\gamma_{x,u}) \coloneqq E_{f^\theta q^\theta}\left[\log\left(\frac{d\rho^\theta_{x,u}}{d\rho^\gamma_{x,u}}\right)\right],
	$$
	given $\rho^\theta_{x,u}$ is absolutely continuous with respect to $\rho^\gamma_{x,u}$.
\end{definition}

Note that the expectation operator $E_{f^\theta q^\theta}[\cdot\mid x,u]$ was defined earlier in Section \ref{ProblemFormulation}, such that we integrate over the random reward and next state, for one step only. 
Next, we specify the decision rule to characterize the TS policy.

\subsection{Decision Rule\label{DecisionRule}} 

\begin{definition}
	The Thompson sampling policy $\tau = \{\tau_t\}$ is a sequence of stochastic kernels $\tau_t$ on $\mathcal{U}$ given $h_t$ and $\theta_t$, satisfying 
	\begin{equation*}
	\tau_t(\cdot\mid h_t,\theta_t) \coloneqq \mu^{\theta_t}(\cdot\mid x_t).
	\end{equation*}
\end{definition}

Under Assumption \ref{uniqueSoln}, $\mu^{\theta_t}$ is a stationary, deterministic, and Markovian policy. Since $\mu^{\theta_t}$ only depends on $x_t$, (by definition) the $\tau$ policy depends on $x_t$, instead of $h_t$. In each period $t$, the TS decision rule samples $\theta_t$ and employs $\mu^{\theta_t}$, i.e., it picks the control that maximizes the expected infinite-horizon discounted reward by treating $\theta_t$ as the true value of the unknown parameter $\theta$. The TS decision rule is deterministic, given $\theta_t$. However, until the (random and history-dependent) sample $\Theta_t$ is drawn from the posterior distribution $\pi_t(\theta_t\mid H_t)$, it is a randomized decision rule. 

Recall the evolution of the stochastic process from Section \ref{overview}. Based on the state-control pair, the DM observes a noisy reward generated by $f^\theta(\cdot\mid x_t,u_t)$, and thus, cannot immediately identify the true value $\theta$. Then, through the transition density $q^\theta(\cdot\mid x_t,u_t)$, the current state transitions into the next state. After observing the reward and the transition, the DM updates the posterior on every $\gamma\in\mathcal{P}$ using (\ref{BayesTheorem}). Afterwards, a new parameter estimate $\theta_{t+1}$ is sampled from the updated distribution, leading to the next control $u_{t+1}$. 

We underline that the history vector includes the sample, i.e., $H_t$ contains the period-$t$ sample $\theta_t$, which renders the TS policy well defined on the probability space $(\Omega, \mathcal{B}(\Omega), \mathbb{P}_{x_0}^{\tau,\theta})$, defined in Section \ref{ProblemFormulation}.

The following lemma formalizes the intuitive observation that TS reduces to the $\theta$-optimal policy when there is no parameter uncertainty.

\begin{lemma}[Degenerate prior] \label{degeneratePrior}
	TS is equivalent to the $\theta$-optimal policy when the prior distribution is degenerate, i.e., $\pi_0(\theta\mid h_0)=1$.
\end{lemma}

\proof[Proof of Lemma \ref{degeneratePrior}]
When the Bayesian update is conducted with a degenerate distribution, it returns a degenerate distribution. We show this by induction. We have $\pi_0(\theta\mid h_0)=1$ and assume $\pi_n(\theta\mid H_n)=1$.
\begin{align*}
\pi_{n+1}(\theta\mid H_{n+1}) \coloneqq& \frac{\mathcal{L}^\theta(H_{n+1})\pi_0(\theta\mid h_0)}{\sum_{\gamma\in\mathcal{P}}\mathcal{L}^\gamma(H_{n+1})\pi_0(\gamma\mid h_0)} \\
=& \frac{\mathcal{L}^\theta(H_{n+1})\pi_0(\theta\mid h_0)}{\mathcal{L}^\theta(H_{n+1})\pi_0(\theta\mid h_0) + \sum_{\gamma\ne\theta\in\mathcal{P}}\mathcal{L}^\gamma(H_{n+1})\pi_0(\gamma\mid h_0)} = \frac{\mathcal{L}^\theta(H_{n+1})}{\mathcal{L}^\theta(H_{n+1})} = 1.
\end{align*} 

Since the posterior distribution is degenerate, TS always samples the true parameter $\theta$ from the parameter space. Therefore, the DM who runs the $\tau$ policy ends up implementing the $\theta$-optimal policy.
\endproof

\section{Expected Residual Regret Bounds} \label{regretBound}
The goal of this section is to show that the expected residual regret of TS vanishes (i.e., converges to 0) in an exponential rate. To do so, we combine the machinery of \cite{hernandez2012adaptive} with \cite{kim2017thompson}. In particular, we bound the expected residual regret by bounding $\mathbb{E}_{x_0}^{\tau,\theta}[\phi^\theta(X_t,U_t)\mid \theta_t\ne\theta]$ and $\mathbb{E}_{x_0}^{\tau,\theta}[\phi^\theta(X_t,U_t)]$. 

The DM draws a sample in each period $t$, represented by the random variable $\Theta_t$. Recall from Section \ref{PosteriorDistribution}, given the true parameter $\theta\in\mathcal{P}$, 
\begin{equation} \label{posterior}
\pi_t(\theta\mid H_t) = \mathbb{P}^{\tau,\theta}_{x_0}(\Theta_t = \theta\mid H_t)
\end{equation}
is the probability that the sample $\Theta_t$ is equal to $\theta$, conditional on $H_t$. Since the condition is not a known $h_t$, (\ref{posterior}) is also a random variable. Similarly, 
\begin{equation} \label{posteriorNot}
1-\pi_t(\theta\mid H_t) = \mathbb{P}_{x_0}^{\tau,\theta}(\Theta_t\ne\theta\mid H_t)
\end{equation}
is the (random) probability that the sample $\Theta_t$ is \emph{not} equal to $\theta$, given (random) $H_t$. Taking the expectation of both sides of (\ref{posteriorNot}),
\begin{equation} \label{posteriorExp}
\mathbb{E}_{x_0}^{\tau,\theta}[1-\pi_t(\theta\mid H_t)] = \mathbb{E}_{x_0}^{\tau,\theta} [\mathbb{P}_{x_0}^{\tau,\theta}(\Theta_t\ne\theta\mid H_t)],
\end{equation}
resolves the uncertainty of $H_t$. By the law of iterated expectations, (\ref{posteriorExp}) simplifies into
\begin{align} 
1-\pi_t(\theta) = \mathbb{P}_{x_0}^{\tau,\theta}(\Theta_t\ne\theta),
\end{align}
i.e., the posterior probability that the sample is \emph{not} equal to the true parameter $\theta$. Then, by the law of total expectation, we partition the expectation of $\phi^\theta$, i.e.,
\begin{align*}
\mathbb{E}_{x_0}^{\tau,\theta}[\phi^\theta(X_t,U_t)] = \mathbb{P}_{x_0}^{\tau,\theta}(\Theta_t\ne \theta) &\mathbb{E}_{x_0}^{\tau,\theta}[\phi^\theta(X_t,U_t)\mid \Theta_t\ne \theta]\\
+ \mathbb{P}_{x_0}^{\tau,\theta}(\Theta_t = \theta) &\mathbb{E}_{x_0}^{\tau,\theta}[\phi^\theta(X_t,U_t)\mid \Theta_t = \theta],
\end{align*}
where $U_t$ is the control that maximizes the reward-to-go by treating the sampled estimate $\Theta_t$ as the true value of the unknown parameter $\theta$. We can simply rewrite the probability terms to obtain,
\begin{align*} 
\mathbb{E}_{x_0}^{\tau,\theta}[\phi^\theta(X_t,U_t)] = (1-\pi_t(\theta))  &\mathbb{E}_{x_0}^{\tau,\theta}[\phi^\theta(X_t,U_t)\mid \Theta_t\ne \theta] \\ 
+\pi_t(\theta)&\mathbb{E}_{x_0}^{\tau,\theta}[\phi^\theta(X_t,U_t)\mid \Theta_t = \theta]. 
\end{align*}

But then, when $\Theta_t=\theta$, by the decision rule $\tau_t(U_t\mid h_t,\theta)$, the optimal control (of state $x_t$) is taken, and the temporal difference error $\phi^\theta(X_t,U_t)$ becomes 0, by definition. Hence, when $\Theta_t=\theta$, 
\begin{equation} \label{expectationPhi}
\mathbb{E}_{x_0}^{\tau,\theta}[\phi^\theta(X_t,U_t)] = (1-\pi_t(\theta)) \mathbb{E}_{x_0}^{\tau,\theta}[\phi^\theta(X_t,U_t)\mid \Theta_t\ne \theta].
\end{equation}

Note that, although we express the $\tau$ policy as a sequence of stochastic kernels, it is a deterministic policy by Assumption \ref{uniqueSoln}, when conditioned on $\theta_t$.
Before presenting the main result of this section, we first bound the expectation of $\phi^\theta(X_t,U_t)$.

\begin{lemma}[Lower bound on expected $\phi^\theta$] \label{conditionalPhi} 
	The expected value of $\phi^\theta(X_t,U_t)$, conditional on $\Theta_t\ne \theta$, is lower bounded by a non-positive constant, i..e,
	\begin{equation} \label{conditional exp phi}
	\mathbb{E}_{x_0}^{\tau,\theta}[\phi^\theta(X_t,U_t)\mid \Theta_t\ne \theta]\ge -2M\left(\frac{1+\beta}{1-\beta}\right),
	\end{equation}
	where $M\ge 0$ is the upper bound on the absolute value of the expected reward, per Remark \ref{boundedExpReward}.
\end{lemma} 
\proof[Proof of Lemma \ref{conditionalPhi}]
In this proof, we slightly modify the $U_t$ notation to indicate its dependence on the sample. Let $U_{t}^{\Theta_t}$ denote the control picked when sample $\Theta_t$ is drawn from $\mathcal{P}$, and let $U_{t}^{\theta}$ represent the control picked when knowing $\theta$. Otherwise, we cannot distinguish the $U_t$ controls.
\begin{align*}
\mathbb{E}_{x_0}^{\tau,\theta}[\phi^\theta(X_t,U_{t}^{\Theta_t})\mid \Theta_t\ne \theta] 
= & \mathbb{E}_{x_0}^{\tau,\theta}[R_t(X_t,U_{t}^{\Theta_t})+\beta\int \nu^\theta(x_{t+1})q^\theta(dx_{t+1}\mid X_t,U_{t}^{\Theta_t}) \, d\eta \mid \Theta_t\ne \theta] \\
&\qquad - \mathbb{E}_{x_0}^{\tau,\theta}[\nu^\theta(X_t)\mid \Theta_t\ne \theta],
\end{align*}
which is equal to
\begin{align*}
& \mathbb{E}_{x_0}^{\tau,\theta}\left[R_t(X_t,U_{t}^{\Theta_t})+\beta\int \nu^\theta(x_{t+1})q^\theta(dx_{t+1}\mid X_t,U_{t}^{\Theta_t}) \, d\eta \mid \Theta_t\ne \theta\right] \\
&\qquad - \mathbb{E}_{x_0}^{\tau,\theta}\left[R_t(X_t,U_{t}^{\theta})+\beta\int \nu^\theta(x_{t+1})q^\theta(dx_{t+1}\mid X_t,U_{t}^{\theta}) \, d\eta \mid \Theta_t\ne \theta\right].
\end{align*}

We can rearrange the above expression to obtain
\begin{align}
&\, \mathbb{E}_{x_0}^{\tau,\theta}[R_t(X_t,U_{t}^{\Theta_t})\mid \Theta_t\ne \theta] - \mathbb{E}_{x_0}^{\tau,\theta}[R_t(X_t,U_{t}^{\theta})\mid \Theta_t\ne \theta] \label{reward difference} \\
&\qquad + \mathbb{E}_{x_0}^{\tau,\theta}\left[\beta\int \nu^\theta(x_{t+1})q^\theta(dx_{t+1}\mid X_t,U_{t}^{\Theta_t}) \, d\eta \mid \Theta_t\ne \theta\right] \label{algorithm} \\ 
&\qquad - \mathbb{E}_{x_0}^{\tau,\theta}\left[\beta\int \nu^\theta(x_{t+1})q^\theta(dx_{t+1}\mid X_t,U_{t}^{\theta}) \, d\eta \mid \Theta_t\ne \theta\right] \label{oracle}. 
\end{align}

Then, we introduce the function $g^{\theta}:\mathcal{X}\times\mathcal{P}\to\mathbb{R}_+$
\begin{equation} \label{gFunc}
g^{\theta}(x,\theta) \coloneqq \int \nu^\theta(y)q^\theta(dy\mid x,U_{t}^{\theta}) \, d\eta,
\end{equation}
such that (\ref{algorithm}) is 
$$
\beta\mathbb{E}_{x_0}^{\tau,\theta}[g^{\theta}(X_t,\Theta_t)\mid \Theta_t\ne \theta]
$$
and (\ref{oracle}) is
$$
-\beta\mathbb{E}_{x_0}^{\tau,\theta}[g^{\theta}(X_t,\theta)\mid \Theta_t\ne \theta].
$$ 

By Remark \ref{boundedExpReward} and the infinite series property, we have
$$
\left\vert \nu^\theta(x)\right\vert \le \frac{M}{1-\beta}, \quad \forall x\in\mathcal{X}.
$$

By (\ref{gFunc}), 
$$
\left\vert g^{\theta}(x,\theta)\right\vert \le \frac{M}{1-\beta}\int q^\theta(dy\mid x,U_{t}^{\theta}) \, d\eta \le \frac{M}{1-\beta}.
$$

Taking the conditional expectation yields the same upper and lower bounds, i.e.,
$$
\left\vert \mathbb{E}_{x_0}^{\tau,\theta}[g^{\theta}(x,\theta)\mid \Theta_t\ne \theta]\right\vert \le \frac{M}{1-\beta}.
$$

We lower bound (\ref{algorithm})
$$
\beta\mathbb{E}_{x_0}^{\tau,\theta}[g^{\theta}(X_t,\Theta_t)\mid \Theta_t\ne \theta] \ge \frac{-\beta M}{1-\beta}, 
$$
and upper bound (\ref{oracle})
$$
\beta\mathbb{E}_{x_0}^{\tau,\theta}[g^{\theta}(X_t,\theta)\mid \Theta_t\ne \theta] \le \frac{\beta M}{1-\beta}, 
$$
to get a lower bound on their difference,
$$
\beta\mathbb{E}_{x_0}^{\tau,\theta}[g^{\theta}(X_t,\Theta_t)\mid \Theta_t\ne \theta] - \beta\mathbb{E}_{x_0}^{\tau,\theta}[g^{\theta}(X_t,\theta)\mid \Theta_t\ne \theta] \ge \frac{-2\beta M}{1-\beta}.
$$

The difference between the first two reward terms, i.e., (\ref{reward difference}), can be no smaller than $-2M$, thus
\begin{align*}
\mathbb{E}_{x_0}^{\tau,\theta}[\phi^\theta(X_t,U_t^{\Theta_t})\mid \Theta_t\ne \theta] \ge -2M + \frac{2\beta M}{1-\beta} = -2M\left(\frac{1+\beta}{1-\beta}\right).
\end{align*}
\endproof




Lemma 4 of \cite{kim2017thompson} bounds the expected probability of not selecting $\theta$ at time $t$ under the average-reward criterion and finite state and action spaces. We extend this result to the discounted infinite-horizon setting with Borel (possibly infinite) state and action spaces.  To do so, we introduce the following assumption, which generalizes the conditions in \cite{kim2017thompson} to our setting.

\begin{assumption} \label{relativeEntropy}
	We assume 
	\begin{align} \label{f_bounded_away}
	\inf\limits_{x\in\mathcal{X},u\in\mathcal{U},r\in\mathscr{R}_c} f^{\gamma}(r \mid x,u) > 0, \quad \forall \gamma\in\mathcal{P}
	\end{align}
	and	
	\begin{align} \label{q_bounded_away}
	\inf_{x\in\mathcal{X},u\in\mathcal{U},y\in\mathcal{X}} q^{\gamma}(y \mid x,u) > 0, \quad \forall \gamma\in\mathcal{P}.
	\end{align}
	
	For any $x\in\mathcal{X}$, $u\in\mathcal{U}$, and any distinct parameter value, $\gamma\ne\theta \in\mathcal{P}$, there exists a positive constant $\epsilon(x,u,\theta,\gamma) > 0$ such that
	\begin{align} \label{entropy_bounded_away}
	\inf_{x\in\mathcal{X},u\in\mathcal{U}} \mathcal{K}(\rho_{x,u}^{\theta}\mid\rho_{x,u}^{\gamma}) > \epsilon(x,u,\theta,\gamma).
	\end{align}
\end{assumption}  

Assumption \ref{relativeEntropy} ensures that different parameter values remain statistically distinguishable. More specifically, it ensures that given $\theta\ne \gamma$, the probability measures $\rho^\theta_{x,u}$ and $\rho^\gamma_{x,u}$ are distinguishable as measured by the relative entropy. In Appendix \ref{AppendixAssumptionVerif}, we illustrate the implications of Assumption \ref{relativeEntropy} through an example. 

\textcolor{black}{This assumption has two important consequences. First, it ensures accessibility across the state space and rules out absorbing regions. Second,} it implies a form of \emph{passive learning}, in the sense that observations are informative about the underlying parameter regardless of the actions taken. Such settings are typically associated with slow learning dynamics, as information is not actively directed toward resolving uncertainty. In contrast, we show that the expected residual regret of TS converges to zero at an exponential rate. In addition, this property is consistent with the conditions used in \cite{kim2017thompson}, and allows us to extend their results to a more general setting with Borel state and action spaces without imposing additional structural assumptions on the underlying dynamics.
The following lemma extends Lemma 4 of \cite{kim2017thompson} to our setting. 

\begin{lemma}[Extension of Lemma 4 of \cite{kim2017thompson}] \label{KimLemma} 
	Suppose Assumption \ref{relativeEntropy} holds. If TS is implemented from any initial state $x_0\in\mathcal{X}$, then there exist
constants $a_\theta$, $b_\theta > 0$ such that 
	\begin{align} \label{post_sampling_error}
	\mathbb{E}_{x_0}^{\tau,\theta}\left[1-\pi_t(\theta \mid H_t)\right]\le a_\theta e^{-b_\theta t},
	\end{align}
	where $a_\theta$ and $b_\theta$ are defined as in \cite{kim2017thompson}.
\end{lemma}

\proof[Proof of Lemma \ref{KimLemma}]
Because Lemma \ref{KimLemma} is an adaptation of Lemma 4 of \cite{kim2017thompson} into our setting, we first walk the reader through the proof of \cite{kim2017thompson} while using our notation.
The proof initially defines the stochastic process, for any $\gamma\ne\theta$,
$$
Z_t^\gamma = \sum_{s=0}^t \log\Lambda_s^\gamma,
$$
where, using our notation, 
\begin{align*}
\Lambda_0^\gamma &= 1 \\
\Lambda_s^\gamma &= \frac{f^\theta(R_{s-1}\mid X_{s-1},U_{s-1}) q^\theta(X_s\mid X_{s-1},U_{s-1})}{f^\gamma(R_{s-1}\mid X_{s-1},U_{s-1}) q^\gamma(X_s\mid X_{s-1},U_{s-1})},
\end{align*} 
for $0<s\le t$.	
Next, \cite{kim2017thompson} defines filtration $(\mathcal{H}_t : t\ge 0)$ by $\mathcal{H}_t = \sigma(H_t)$, where $H_t$ is by the definition in (\ref{history_vector}). Then, it follows that the stochastic process $Z_t^\gamma$ is a submartingale with respect to $\mathcal{H}_t$ under probability measure $\mathbb{P}_{x_0}^{\tau,\theta}$. It is crucial that $Z_t^\gamma$ is a submartingale. \cite{kim2017thompson} decomposes it into an $\mathcal{H}_t$ martingale under $\mathbb{P}_{x_0}^{\tau,\theta}$,
$$
M_t^\gamma \coloneqq \sum_{s=0}^t ( \log \Lambda^\gamma_s - \mathbb{E}^{\tau,\theta}_{x_0}[\log \Lambda^\gamma_s \mid \mathcal{H}_{s-1}] ),
$$
and a predictable process,
$$
A_t^\gamma \coloneqq \sum_{s=0}^t \mathbb{E}^{\tau,\theta}_{x_0}[\log \Lambda^\gamma_s \mid \mathcal{H}_{s-1}].
$$

We explain how the proof extends to our setting through Assumption \ref{relativeEntropy}.
The argument is twofold. 
The first requirement is that the increments of $M_t^\gamma$ are bounded above and below by some $d$, i.e., 
\begin{align} \label{boundedInc}
\vert \log \Lambda^\gamma_s - \mathbb{E}^{\tau,\theta}_{x_0}[\log \Lambda^\gamma_s \mid \mathcal{H}_{s-1}] \vert \le d,
\end{align}
for some $d>0$. 
To see that (\ref{boundedInc}) holds in our setting, note that 
by (\ref{f_bounded_away}) and (\ref{q_bounded_away}), i.e., $f^\gamma(r \mid x,u)$ and $q^\gamma(y \mid x,u)$ are bounded away from 0, we have that $\vert \log f^\gamma(r \mid x,u) \vert < \infty$ and $\vert \log q^\gamma(y \mid x,u)\vert < \infty$. This satisfies (\ref{boundedInc}), which is needed for Azuma's inequality, a crucial step of \cite{kim2017thompson}'s proof, to hold. 
The second requirement of the adaptation of the proof is
\begin{align} \label{increasingProc}
\sum\limits_{s=0}^t \mathbb{E}_{x_0}^{\tau,\theta}[\log \Lambda^\gamma_s \mid \mathcal{H}_{s-1}] \ge \epsilon t,
\end{align}
i.e., (\ref{increasingProc}) is an increasing predictable process.
By (\ref{entropy_bounded_away}), it follows that
\begin{align} \label{eps_lower_bounds}
\mathbb{E}^{\tau,\theta}_{x_0}[\log \Lambda^\gamma_s \mid \mathcal{H}_{s-1}] > \epsilon(x,u,\theta,\gamma) \coloneqq \epsilon, \quad \forall s\le t,
\end{align}
i.e., each increment of $A_t^\gamma$ is strictly positive, which satisfies (\ref{increasingProc}). The reader is referred to \cite{kim2017thompson} for the details of why (\ref{eps_lower_bounds}) holds.
\endproof

By the law of iterated expectations, (\ref{post_sampling_error}) simplifies into 
\begin{equation} \label{KimSimplifed}
1-\pi_t(\theta) \le a_\theta e^{-b_\theta t}.
\end{equation}

We are now ready to introduce the main result of this section.
\begin{proposition}[Upper bound on $\mathcal{R}_{x_0}^{\tau,\theta}(n)$] 
	\label{proposition1} When Assumption \ref{relativeEntropy} holds, the expected residual regret converges to 0 exponentially fast. In particular,
	$$
	\mathcal{R}_{x_0}^{\tau,\theta}(n) \coloneqq \mathbb{E}_{x_0}^{\tau,\theta}[\nu^\theta(X_n)] - V_{x_0}^{\tau,\theta}(n) \,\, \le \,\, \frac{2M(1+\beta)a_\theta e^{-b_\theta n}}{(1-\beta)^2},
	$$
	where $a_\theta$ and $b_\theta$ are positive constants, defined in Lemma 4 of \cite{kim2017thompson}.
\end{proposition}

\proof[Proof of Proposition \ref{proposition1}]
Equations (\ref{expectationPhi}), (\ref{conditional exp phi}) and (\ref{KimSimplifed}) together yield
\begin{equation} \label{negPhi}
-\mathbb{E}_{x_0}^{\tau,\theta}[\phi^\theta(X_t,U_t)] \le
\frac{2M(1+\beta)a_\theta e^{-b_\theta t}}{1-\beta}.
\end{equation}

By the proof of Lemma \ref{phi}, $\phi^\theta(X_t,U_t)$ and $\mathcal{R}_{x_0}^{\tau,\theta}(n)$ are related, see (\ref{phi_residualregret}). Then,
by (\ref{negPhi}), the right-hand side of (\ref{phi_residualregret}) is upper bounded,
$$
- \sum_{t=n}^\infty \beta^{t-n} \mathbb{E}_x^{\tau,\theta}[\phi^\theta(X_t,U_t)] \le \sum_{t=n}^\infty \beta^{t-n} \frac{2M(1+\beta)a_\theta e^{-b_\theta t}}{1-\beta}.
$$

Since $b_\theta \ge 0$, we have $e^{-b_\theta t}\le e^{-b_\theta n}$ for all $t\ge n$. Thus, the above equation becomes
\begin{equation} \label{infinite_geom}
- \sum_{t=n}^\infty \beta^{t-n} \mathbb{E}_x^{\tau,\theta}[\phi^\theta(X_t,U_t)] \le \sum_{t=n}^\infty \beta^{t-n} \frac{2M(1+\beta)a_\theta e^{-b_\theta n}}{1-\beta}.
\end{equation}

By (\ref{phi_residualregret}) and applying the infinite geometric series formula to (\ref{infinite_geom}), we obtain 
\begin{align*}
-\sum_{t=n}^\infty \beta^{t-n} \mathbb{E}_{x_0}^{\tau,\theta}[\phi^\theta(X_t,U_t)] &= \mathbb{E}_{x_0}^{\tau,\theta}[\nu^\theta(X_n)] - V_{x_0}^{\tau,\theta}(n) \\
&= \mathcal{R}_{x_0}^{\tau,\theta}(n) \le \frac{2M(1+\beta)a_\theta e^{-b_\theta n}}{(1-\beta)^2},
\end{align*}
thus showing that the upper bound on the expected residual regret decays exponentially.
\endproof

\section{Almost sure learning for Thompson Sampling\label{completeLearning}}

In this section, we strengthen our results on the learning behavior of TS in $\theta$-MDPs. In contrast to the previous section, where we established convergence of the residual regret to 0 in expectation, we now prove almost sure convergence with respect to $\mathbb{P}_{x_0}^{\tau,\theta}$. More specifically, in Section \ref{subsec:completelearning} we show that the posterior distribution under TS converges $\mathbb{P}_{x_0}^{\tau,\theta}$-a.s. to a point mass at $\theta$, a property known as \emph{complete learning}. In Section \ref{sec:probabilisticResidualRegret}, we further show that the probabilistic residual regret converges to zero $\mathbb{P}_{x_0}^{\tau,\theta}$-a.s., that is, along almost every sample path.



\subsection{Complete Learning\label{subsec:completelearning}}

In this subsection, we study the asymptotic behavior of the posterior distribution under TS. Previous work by \cite{kim2017thompson} shows that the expected posterior sampling error of TS converges to zero at an exponential rate; we extend this result to our setting in Lemma \ref{KimLemma}. Building on it, we strengthen the convergence from expectation to almost sure convergence. To this end, we impose the following assumption, which ensures that the posterior admits a well-defined limit.


\begin{assumption}[Existence of the limit of $\pi_t(\theta\mid H_t)$]
	\label{existenceofLimitPosterior}
	Suppose that $\lim\limits_{t\to\infty} \pi_t(\theta\mid H_t)$ exists $\mathbb{P}_{x_0}^{\tau,\theta}$-a.s.
\end{assumption}

This assumption rules out pathological oscillatory behavior of the posterior sequence and allows us to characterize its limiting behavior.
Under Assumption \ref{relativeEntropy}, the expected posterior sampling error of TS vanishes. We will show that this, i.e., 
$\lim\limits_{t\to\infty} \mathbb{E}_{x_0}^{\tau,\theta}[\pi_t(\theta\mid H_t)] = 1$, 
and Assumption \ref{existenceofLimitPosterior} together imply
\begin{equation}\label{eq:completelearning}
\lim_{t\to\infty} \pi_t(\theta\mid H_t) = 1, \quad \mathbb{P}_{x_0}^{\tau,\theta}\textnormal{-almost surely}.
\end{equation}

The next lemma is the final piece to show the occurrence of complete learning.

\begin{lemma} \label{contradictionLemma}
	Suppose that TS does not exhibit complete learning, i.e., $\lim\limits_{t\to\infty} \pi_t(\theta\mid H_t) < 1$, $\mathbb{P}_{x_0}^{\tau,\theta}$-a.s. Then, there exists an $n\ge 1$ s.t. 
	$$
	\mathbb{P}_{x_0}^{\tau,\theta}\left(\lim_{t\to\infty} \pi_t(\theta\mid H_t) < 1-\frac{1}{n}\right) > 0, \quad \mathbb{P}_{x_0}^{\tau,\theta}\textnormal{-almost surely.}
	$$
\end{lemma}

\proof[Proof of Lemma \ref{contradictionLemma}]
By the definition of almost-sure convergence, we have
$$
\mathbb{P}_{x_0}^{\tau,\theta}\left(\lim_{t\to\infty} \pi_t(\theta\mid H_t) < 1\right) = 1.
$$

Therefore,
\begin{align*}
1 = \mathbb{P}_{x_0}^{\tau,\theta}\left(\lim_{t\to\infty} \pi_t(\theta\mid H_t) < 1\right) &=\mathbb{P}_{x_0}^{\tau,\theta}\left(\bigcup_{n=1}^\infty\left\{\lim_{t\to\infty} \pi_t(\theta\mid H_t)<1-\frac{1}{n}\right\}\right) \\
&\le \sum_{n=1}^\infty \mathbb{P}_{x_0}^{\tau,\theta}\left(\lim_{t\to\infty} \pi_t(\theta\mid H_t) < 1 - \frac{1}{n}\right). 
\end{align*}
where the inequality is by Boole's inequality. Hence, Lemma \ref{contradictionLemma} is verified.
\endproof

We are now equipped to show the main result of this section.

\begin{theorem}[TS learns $\theta$ as $t\to\infty$] \label{CLtheoremSECOND} Suppose that Assumptions \ref{relativeEntropy} and \ref{existenceofLimitPosterior} hold. Then \eqref{eq:completelearning} holds.
\end{theorem}

\proof[Proof of Theorem \ref{CLtheoremSECOND}]
By Lemma \ref{KimLemma}, Assumption \ref{existenceofLimitPosterior}, and the Dominated Convergence Theorem, it follows that
$$
1 = \lim_{t\to\infty}\mathbb{E}_{x_0}^{\tau,\theta}[\pi_t(\theta\mid H_t)] = \mathbb{E}_{x_0}^{\tau,\theta}\left[\lim_{t\to\infty} \pi_t(\theta\mid H_t)\right], \quad \mathbb{P}_{x_0}^{\tau,\theta}\textnormal{-a.s.}
$$

Then,
\begin{align*}
1&= \mathbb{E}_{x_0}^{\tau,\theta}\left[\lim_{t\to\infty} \pi_t(\theta\mid H_t)\right] \\
&= \mathbb{E}_{x_0}^{\tau,\theta}\left[\lim_{t\to\infty} \pi_t(\theta\mid H_t)\Big\vert \left\{\lim_{t\to\infty} \pi_t(\theta\mid H_t) < 1-\frac{1}{n}\right\}\right] \mathbb{P}_{x_0}^{\tau,\theta}\left(\lim_{t\to\infty} \pi_t(\theta\mid H_t) < 1-\frac{1}{n}\right) \\
&\qquad + \mathbb{E}_{x_0}^{\tau,\theta}\left[\lim_{t\to\infty} \pi_t(\theta\mid H_t)\Big\vert \left\{\lim_{t\to\infty} \pi_t(\theta\mid H_t) \ge 1-\frac{1}{n}\right\}\right] \mathbb{P}_{x_0}^{\tau,\theta}\left(\lim_{t\to\infty} \pi_t(\theta\mid H_t) \ge 1-\frac{1}{n}\right) \\
&\le  (1-1/n)\mathbb{P}_{x_0}^{\tau,\theta}\left(\lim_{t\to\infty} \pi_t(\theta\mid H_t) < 1-\frac{1}{n}\right) + 1-\mathbb{P}_{x_0}^{\tau,\theta}\left(\lim_{t\to\infty} \pi_t(\theta\mid H_t) < 1-\frac{1}{n}\right) \\
&= 1 - \frac{1}{n}\mathbb{P}_{x_0}^{\tau,\theta}\left(\lim_{t\to\infty} \pi_t(\theta\mid H_t) < 1-\frac{1}{n}\right), \quad \mathbb{P}_{x_0}^{\tau,\theta}\textnormal{-a.s.}
\end{align*}

However, by Lemma \ref{contradictionLemma}, $\lim\limits_{t\to\infty} \pi_t(\theta\mid H_t) < 1$ implies
\begin{align*}
1 &= \mathbb{E}_{x_0}^{\tau,\theta}\left[\lim_{t\to\infty} \pi_t(\theta\mid H_t)\right] 
\\
&\le 1 - \frac{1}{n}\mathbb{P}_{x_0}^{\tau,\theta}\left(\lim_{t\to\infty} \pi_t(\theta\mid H_t) < 1-\frac{1}{n}\right) < 1, \quad \mathbb{P}_{x_0}^{\tau,\theta}\textnormal{-almost surely,} 
\end{align*}
leading to contradiction. For all $n\ge 1$, it must be that
$$
\mathbb{P}_{x_0}^{\tau,\theta}\left(\lim_{t\to\infty} \pi_t(\theta\mid H_t) < 1-\frac{1}{n}\right) = 0, \quad \mathbb{P}_{x_0}^{\tau,\theta}\textnormal{-almost surely.}
$$

Thus, we must have $\lim\limits_{t\to\infty} \pi_t(\theta\mid H_t) = 1$, $\mathbb{P}_{x_0}^{\tau,\theta}$-almost surely. This completes the proof.
\endproof

\subsection{Vanishing Probabilistic Residual Regret\label{sec:probabilisticResidualRegret}}
We now study the asymptotic behaviour of the residual regret. To do so, we assume that its limit exists almost surely under $\mathbb{P}_{x_0}^{\tau,\theta}$.

\begin{assumption}[Existence of the limit of $\mathbb{R}_{x_0}^{\tau,\theta}(n)$] \label{existenceofLimitRR}
	Suppose that $\lim\limits_{n\to\infty} \mathbb{R}_{x_0}^{\tau,\theta}(n)$ exists $\mathbb{P}_{x_0}^{\tau,\theta}$-a.s.
\end{assumption}

From (\ref{resRegDiff1}) observe that the residual regret is the difference of two terms. Consider the first term $\nu^\theta(X_n)$. Under some technical conditions, if the limit of the state process $\{X_n\}$ exists, then the limit of $\nu^\theta(X_n)$ exists. However, since we do not impose any restrictions on the underlying chain structure and the underlying stochastic process is history-dependent, we cannot guarantee that $\lim\limits_{n\to\infty} X_n$ exists, and thus, there could be cases where $\lim\limits_{n\to\infty} \nu^\theta(X_n)$ does not converge\footnote{We do not require aperiodicity; the chain could be periodic, leading to oscillating rewards.}. Therefore, we directly assume that the limit of the residual regret exists. 
Similar to the expected residual regret, the following corollary shows that under certain conditions the probabilistic residual regret also vanishes in the limit.

\begin{corollary} \label{corollary1} When the conditions of Proposition \ref{proposition1} and Assumption \ref{existenceofLimitRR} hold, the residual regret of TS vanishes, i.e.,
	$$
	\lim\limits_{n\to\infty} \mathbb{R}_{x_0}^{\tau,\theta}(n) = 0, \quad \mathbb{P}_{x_0}^{\tau,\theta}\textnormal{-almost surely}.
	$$
\end{corollary}

\proof[Proof of Corollary \ref{corollary1}]
Recall that the probabilistic residual regret converges $\mathbb{P}_{x_0}^{\tau,\theta}\textnormal{-a.s.}$ to 0 if there exists a null set  $\mathbf{A}\in\mathcal{B}(\Omega)$ with $\mathbb{P}_{x_0}^{\tau,\theta}(\mathbf{A})=0$ such that the statement holds if $\omega\notin\mathbf{A}$. Given Assumption \ref{existenceofLimitRR}, it suffices to show that $\forall$ $\epsilon > 0$, $\nexists$ $\mathbf{A}\in\mathcal{B}(\Omega)$ with $\mathbb{P}_{x_0}^{\tau,\theta}(\mathbf{A})=\epsilon > 0$ such that $\lim\limits_{n\to\infty}\mathbb{R}_{x_0}^{\tau,\theta}(n)\ne 0$ $\forall \, \omega\in\mathbf{A}$. We prove this by contradiction. 

Suppose that $\exists$ $\mathbf{A}\in\mathcal{B}(\Omega)$ with $\mathbb{P}_{x_0}^{\tau,\theta}(\mathbf{A}) = \epsilon^\prime > 0$ for some $\epsilon^\prime>0$ such that $\lim\limits_{n\to\infty}\mathbb{R}_{x_0}^{\tau,\theta}(n)\ne 0$ $\forall \, \omega\in\mathbf{A}$. Since $\mathbb{R}_{x_0}^{\tau,\theta}(n) > 0 $ by construction, this limit is strictly positive, i.e., $\lim\limits_{n\to\infty}\mathbb{R}_{x_0}^{\tau,\theta}(n) > 0$ $\forall \, \omega\in\mathbf{A}$. 

Then, for any $\delta>0$, we define $\mathbf{A}_\delta$ as the largest measurable subset of $\mathbf{A}$ such that $\mathbf{A}_\delta \subset \{\omega\in\mathcal{B}(\Omega): \lim\limits_{n\to\infty} \mathbb{R}_{x_0}^{\tau,\theta}(n) > \delta \}$. According to Assumption \ref{existenceofLimitRR} and the supposition in the previous paragraph, $\exists$ $\delta^\prime$ such that $\mathbb{P}_{x_0}^{\tau,\theta}(\mathbf{A}_{\delta^\prime})=\epsilon^\prime > 0$. Then, we can write
\begin{align}
\lim\limits_{n\to\infty} \mathcal{R}_{x_0}^{\tau,\theta}(n) &= \lim\limits_{n\to\infty} \mathbb{E}_{x_0}^{\tau,\theta}[\mathbb{R}_{x_0}^{\tau,\theta}(n)] \\
&= \lim\limits_{n\to\infty} \left \{\int_{\mathbf{A}_{\delta^\prime}} \mathbb{R}_{x_0}^{\tau,\theta}(n) \, d\mathbb{P}_{x_0}^{\tau,\theta}(\omega) + \int_{\mathcal{B}(\Omega)\setminus \mathbf{A}_{\delta^\prime}}  \mathbb{R}_{x_0}^{\tau,\theta}(n) \, d\mathbb{P}_{x_0}^{\tau,\theta}(\omega) \right\} \label{limitofInt} \\ 
&= \int_{\mathbf{A}_{\delta^\prime}} \lim\limits_{n\to\infty}  \mathbb{R}_{x_0}^{\tau,\theta}(n) \, d\mathbb{P}_{x_0}^{\tau,\theta}(\omega) +  \int_{\mathcal{B}(\Omega)\setminus \mathbf{A}_{\delta^\prime}} \lim\limits_{n\to\infty}  \mathbb{R}_{x_0}^{\tau,\theta}(n) \, d\mathbb{P}_{x_0}^{\tau,\theta}(\omega) \label{intofLimit} \\
& > \delta^\prime\epsilon^\prime \label{deltaEpsilon} > 0,
\end{align}
which contradicts Proposition \ref{proposition1}.
By construction, the first term in (\ref{limitofInt}) converges to a number strictly greater than 0, and the second term converges to some non-negative number. Because $\mathbb{R}_{x_0}^{\tau,\theta}(n)$ exists by Assumption \ref{existenceofLimitRR} and is finite\footnote{Since the rewards are finite-valued, in any period, the difference in rewards of the first and second terms is bounded by a constant. By the infinite geometric series property, and assuming $\beta\in [0,1)$, $\mathbb{R}_{x_0}^{\tau,\theta}(n)$ is finite.} 
, by the Dominated Convergence Theorem, we can express (\ref{limitofInt}) as (\ref{intofLimit}),
But then, (\ref{deltaEpsilon}) contradicts with Proposition \ref{proposition1}. This implies that our supposition cannot hold, and thus, the residual regret converges $\mathbb{P}_{x_0}^{\tau,\theta}\text{-a.s.}$ to 0.
\endproof

\section{Concluding Remarks and Discussion}

This paper develops a learning framework for sampling-based algorithms in MDPs with an unknown parameter $\theta$ governing rewards and transitions, which we refer to as $\theta$-MDPs. We consider the discounted infinite-horizon criterion with Borel state and action spaces, extending beyond the finite settings typically studied in the literature. 
We introduce a canonical probability space for sampling-based algorithms in $\theta$-MDPs, in which the sampled parameter is incorporated into the history of the process. This formulation highlights the inherently history-dependent nature of $\theta$-MDPs and facilitates their analysis.

As a performance criterion, we adopt the per-period suboptimality gap used in
discounted-MDP regret analysis \cite{zhou,liu} and specialize it to our
parametrized Bayesian setting. Since this quantity measures, along each sample
path, the remaining loss in future performance from the current period onward,
we refer to it as residual regret. Its expected value provides the link
between discounted-MDP regret analysis and ADO,
a classical notion from the adaptive control literature
\cite{manfred1987estimation,hernandez2012adaptive}. In particular, vanishing
expected residual regret is equivalent to ADO in our setting. The expected residual regret also admits a TD representation
in terms of a Bellman residual, linking our criterion to TD learning
\cite{sutton2018reinforcement}.

We then specialize our analysis to Thompson sampling (TS). \textcolor{black}{By exploiting the connections between expected residual regret, ADO, and TD errors, we combine tools from the adaptive control literature \cite{hernandez2012adaptive} with posterior concentration arguments for TS to establish that the expected residual regret vanishes.}
In particular, we show that the expected residual regret of TS decays to zero exponentially fast in the worst case (Proposition \ref{proposition1}). To establish this result, we extend the assumption in \cite{kim2017thompson}, originally developed for finite state and action spaces under the average reward criterion, to a more general setting with Borel state and action spaces and a discounted infinite-horizon objective.
Under this condition, we show two additional results that strengthen the expectation-based guarantee by showing that learning occurs along
almost every sample path. First, we prove that the posterior sampling error converges to zero almost surely (Theorem \ref{CLtheoremSECOND}). To ensure this convergence, we also impose Assumption \ref{existenceofLimitPosterior}, which rules out oscillatory behavior that would prevent complete learning. Second, we show that the sample-path residual regret also vanishes (Corollary \ref{corollary1}). This result relies on the sufficiency condition stated in Assumption \ref{existenceofLimitRR}, which simply requires that the residual regret admits an almost sure limit. 

Our results hold for chain settings where $q^\gamma(\cdot|x,u)$  is strictly positive $\forall \gamma \in \mathcal{P}$,
which rules out chains with absorbing states. Deriving the performance of the residual regret in broader settings is a potential direction of
further research. 
Another future direction of research is to study the implications of Assumptions \ref{existenceofLimitPosterior} and \ref{existenceofLimitRR} in specific problem contexts. In addition to conditions on the underlying chain structure, complete learning may be a necessary but not sufficient condition for Assumption \ref{existenceofLimitRR} to hold. 

\appendix
\section*{Appendices}

\section{Decomposition of the Standard Expected Regret\label{stdRegret}}



A natural starting point for evaluating the performance of an admissible policy is the notion of expected regret, defined as the difference between the optimal value function (obtained under the policy $\mu^\theta$ that knows the true parameter) and the value achieved by a given policy $\mu$. In the discounted infinite-horizon setting, by definitions \eqref{V(0)} and \eqref{v(x)}, this difference is given by
\begin{align} 
E[\textnormal{Regret}^\theta(0,\infty)] &\coloneqq \nu^\theta(x_0) - V_{x_0}^{\mu,\theta}(0) \nonumber \\
&= \mathbb{E}_{x_0}^{\mu^\theta,\theta} \left[\sum_{t=0}^\infty\beta^t R_t\right] - \mathbb{E}_{x_0}^{\mu,\theta} \left[\sum_{t=0}^\infty\beta^t R_t\right], \label{infiniteHorizon_v2}
\end{align}
where $(0,\infty)$ represents the starting and ending periods, inclusive. The expected infinite-horizon regret $E[\textnormal{Regret}^\theta(0,\infty)]$ is a function of two different expectation operators, so the expectation ``$E$'' represents a label rather than a formal mathematical expression. We emphasize that even though $R_t$ appears in both terms of (\ref{infiniteHorizon_v2}), one of them is driven by the process generated by $\mu^\theta$, while the other one is driven by $\mu$. 

By construction, $E[\textnormal{Regret}^\theta(0,\infty)]$ can be partitioned into two components; a finite component that tallies rewards up until some period $n-1$, and the remainder that goes into infinity, i.e.,
\begin{equation} \label{sumOfTwoSums}
E[\textnormal{Regret}^\theta(0,\infty)] = E[\textnormal{Regret}^\theta(0,n-1)] + E[\textnormal{Regret}^\theta(n,\infty)].
\end{equation}

\paragraph{\bf Expected Infinite-Horizon Regret in Period-$n$ Dollars} We denote this quantity by $E[\textnormal{Regret}^\theta_n(0,\infty)]$, where the subscript $n$ indicates that the regret is evaluated in period-$n$ dollars. To express the expected regret of the $\theta$-MDP in period-$n$ dollars, we scale (\ref{infiniteHorizon_v2}) by $\beta^{-n}$ and operate to obtain
\begin{align} 
E[\textnormal{Regret}^\theta_n(0,\infty)] &\coloneqq (\nu^\theta(x_0) - V_{x_0}^{\mu,\theta}(0))\beta^{-n} \\
&= \left[\mathbb{E}_{x_0}^{\mu^\theta,\theta} \left[\sum_{t=0}^\infty\beta^t R_t\right] - \mathbb{E}_{x_0}^{\mu,\theta} \left[\sum_{t=0}^\infty\beta^t R_t\right]\right]\beta^{-n} \label{futInfRegret} \\
&= \left[\mathbb{E}_{x_0}^{\mu^\theta,\theta} \left[\sum_{t=0}^{n-1}\beta^t R_t \right] - \mathbb{E}_{x_0}^{\mu,\theta} \left [\sum_{t=0}^{n-1}\beta^t R_t \right] \right]\beta^{-n} \label{finiteRegret} \\ 
&\qquad+ \left[ \mathbb{E}_{x_0}^{\mu^\theta,\theta} \left[\sum_{t=n}^\infty\beta^t R_t \right] - \mathbb{E}_{x_0}^{\mu,\theta} \left [\sum_{t=n}^\infty\beta^t R_t \right] \right]\beta^{-n} \label{difference1}.
\end{align}

The term \eqref{finiteRegret} corresponds to the expected finite-horizon regret, expressed in period-$n$ dollars. The term \eqref{difference1} captures the expected regret from period $n$ onward, also evaluated in period-$n$ dollars. We denote these components by $E[\textnormal{Regret}^\theta_n(0,n-1)]$ and $E[\textnormal{Regret}^\theta_n(n,\infty)]$, respectively. This yields the following decomposition, analogous to (\ref{sumOfTwoSums}):
\begin{equation*} 
E[\textnormal{Regret}_n^\theta(0,\infty)] = E[\textnormal{Regret}_n^\theta(0,n-1)] + E[\textnormal{Regret}_n^\theta(n,\infty)].
\end{equation*}

Distributing $\beta^{-n}$ inside, (\ref{difference1}) can alternatively be written as 
\begin{equation} \label{differenceBeyondN}
\mathbb{E}_{x_0}^{\mu^\theta,\theta} \left[\sum_{t=n}^\infty\beta^{t-n} R_t \right] - \mathbb{E}_{x_0}^{\mu,\theta} \left[\sum_{t=n}^\infty\beta^{t-n} R_t \right].
\end{equation}
%
By the first remark in Section \ref{MDP}, the infinite geometric series property, and assuming $\beta\in [0,1)$, (\ref{differenceBeyondN}) is upper bounded by $\frac{2M}{1-\beta}$, which is a constant independent of $n$. 

When $\beta\in[0,1)$, the difference in (\ref{infiniteHorizon_v2}) is the expected value of the regret felt by the DM discounted back to period 0, i.e., in period-$0$ ``dollars''. In addition, there is no difference between (\ref{infiniteHorizon_v2}) and (\ref{futInfRegret}).  Finally, note that when $\beta=1$, $E[\textnormal{Regret}^\theta_n(0,n-1)]$ becomes the traditional expected regret over a finite horizon and (\ref{differenceBeyondN}) may grow to infinity in $n$.


\paragraph{\bf Decomposing the Expected Regret from Period $n$ Onward} We further decompose $E[\textnormal{Regret}_n^\theta(n,\infty)]$. By construction, the second term in (\ref{differenceBeyondN}) is equal to the value function of the $\mu$ policy, i.e., $V_{x_0}^{\mu,\theta}(n)$. Moreover, the first term in (\ref{differenceBeyondN}) is equal to the expectation of the optimal value function with respect to the $\theta$-optimal policy, i.e., 
\begin{equation} \label{shifting}
\mathbb{E}_{x_0}^{\mu^\theta,\theta}[\nu^\theta(X_n)] = \mathbb{E}_{x_0}^{\mu^\theta,\theta} \left[\mathbb{E}_{X_n}^{\mu^\theta,\theta} \left[\sum_{t=0}^\infty\beta^{t} R_t^\prime\right]\right] = \mathbb{E}_{x_0}^{\mu^\theta,\theta} \left[\sum_{t=n}^\infty\beta^{t-n} R_t\right].
\end{equation}

In (\ref{shifting}), $X_n$ is generated by running the $\mu^\theta$ policy starting in period $t=0$ from state $x_0$. 
Conditional on the random ``starting'' state $X_n$, $\{R_0^\prime, R_1^\prime, \dots\}$ is the random reward process generated by the optimal policy $\mu^\theta$.
When unconditioned on $X_n$, $R_t^\prime \sim R_{t+n}$ given starting state $x_0$. 
Adding and subtracting $\mathbb{E}_{x_0}^{\mu,\theta}[\nu^\theta(X_n)]$, and regrouping yields the following decomposition.
\begin{align}
E[\textnormal{Regret}^\theta_n(0,\infty)] &= E[\textnormal{Regret}_n^\theta(0,n-1)] \quad \textnormal{(Expected finite-time regret)} \nonumber\\ 
&\qquad+ \mathbb{E}_{x_0}^{\mu^\theta,\theta} [\nu^\theta(X_n)] - \mathbb{E}_{x_0}^{\mu,\theta} [\nu^\theta(X_n)] \quad \text{(Expected state regret)} \label{statePenalty} \\ 
&\qquad+ \mathbb{E}_{x_0}^{\mu,\theta} [\nu^\theta(X_n)] - V_{x_0}^{\mu,\theta}(n) \quad \textnormal{(Expected residual regret)} \label{residualRegret1}. 
\end{align}

Note that the expected finite-time regret is the same quantity as (\ref{finiteRegret}). Next, we discuss (\ref{statePenalty}) and (\ref{residualRegret1}), which we call \emph{expected state regret} and \emph{expected residual regret}, respectively.

\subsection{Expected State Regret\label{sec:statergret}}

The term \eqref{statePenalty} captures the unavoidable future consequences of reaching a suboptimal state after implementing a sampling-based policy for $n$ periods. We formalize this notion below.

\begin{definition}
The \emph{expected state regret} is defined as
\[
\mathcal{S}_{x_0}^{\mu,\theta}(n)
\coloneqq 
\mathbb{E}_{x_0}^{\mu^\theta,\theta} [\nu^\theta(X_n)] 
- 
\mathbb{E}_{x_0}^{\mu,\theta} [\nu^\theta(X_n)].
\]
It represents the expected forward-looking loss induced by the state reached at period $n$. In particular, it measures the difference between the optimal value of the random state reached at period $n$ when policy $\mu$ is implemented from period $0$ to $n$, and the optimal value of the random state that would be reached if the optimal policy $\mu^\theta$ were implemented over the same time horizon.
\end{definition}



Although it tallies the difference in rewards starting from period $n$ into the infinite future, we only denote the starting period $n$ inside the parenthesis.

\begin{figure}[!htb]
	\begin{center}
		\begin{tikzpicture}[
		roundnode/.style={circle, draw=black, very thick, minimum size=10mm,font=\small},
		squarednode/.style={rectangle, draw=black, thick, minimum size=5mm,font=\small},
		]
		\node[roundnode, label={\small Sample path under policy $\mu^\theta$}] (x0Left) {$x_0$};
		\node[roundnode, label={\small Sample path under policy $\mu$}] (x0Right) [right=5cm of x0Left] {$x_0$};
		\node[roundnode] (x1Left) [below=of x0Left] {$X_1$};
		\node[roundnode] (x1Right) [below=of x0Right] {$X_1$};
		\node[roundnode] (xnLeft) [below=2cm of x1Left] {$X_n$};
		\node[roundnode] (xnRight) [below=2cm of x1Right] {$X_n$};
		\node[squarednode] (totalLeft) [below=0.5cm of xnLeft] {$\mathbb{E}_{x_0}^{\mu^\theta,\theta}[\nu^\theta(X_n)]$};
		\node[squarednode] (totalRight) [below=0.5cm of xnRight] {$\mathbb{E}_{x_0}^{\mu,\theta}[\nu^\theta(X_n)]$};
		\draw[thick,->] (x0Left.south) -- ++ (0,-2em) coordinate (x1Left.north);
		\draw[thick,<->] (x0Left.east) -- node [below,midway] {\small Same starting state} (x0Right.west);
		\draw[thick,->] (x0Right.south) -- ++ (0,-2em) coordinate (x1Right.north);
		\draw[thick,<->] (x1Left.east) -- node [below,midway] {\small Potentially different} (x1Right.west);
		\draw[thick,->] (x1Left.south) -- ++ (0,-2em) coordinate (aux1); 
		\draw[thick,->] (x1Right.south) -- ++ (0,-2em) coordinate (aux2); 
		\draw[thick,<-] (xnLeft.north) -- ++ (0,2em) coordinate (aux3); 
		\draw[thick,<-] (xnRight.north) -- ++ (0,2em) coordinate (aux4); 
		\draw[very thick,dotted] (aux1) -- (aux3);
		\draw[very thick,dotted] (aux2) -- (aux4);
		\draw[thick,<->] (xnLeft.east) -- node[below,midway] {\small Potentially different} (xnRight.west);
		\draw[thick,<-] (xnRight.east) -- node[pos=0.8, above right] {\small Oracle reveals $\theta$} +(0:1);
		\draw[thick,->] (totalLeft.south) -- node[pos=0.8,align=left,right] {\small Keep running\\ \small policy $\mu^\theta$} +(-90:1);
		\draw[thick,->] (totalRight.south) -- node[pos=0.8,align=left,right] {\small After time $n$,\\ \small adopt policy $\mu^\theta$} +(-90:1);
		\end{tikzpicture}
		\caption{Illustration of two different sample paths giving rise to the expected state regret.}\label{stateRegretSP}
	\end{center}
\end{figure}
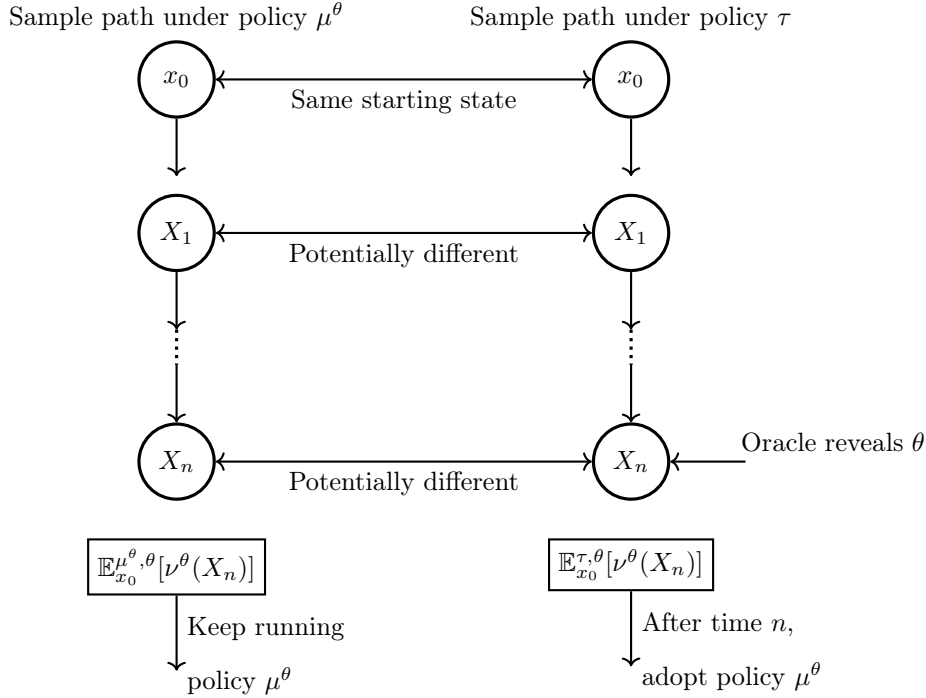

The random state $X_n$ in $\mathbb{E}_{x_0}^{\mu^\theta,\theta} [\nu^\theta(X_n)]$ is induced by running the $\mu^\theta$ policy for $n$ periods, whereas the random $X_n$ in $\mathbb{E}_{x_0}^{\mu,\theta} [\nu^\theta(X_n)]$ arises under the $\mu$ policy. Starting from the respective random states $X_n$, both sample paths follow the $\theta$-optimal policy. Figure \ref{stateRegretSP} illustrates two representative sample paths. If there is no state process or if the state process is iid, then these two terms would be equal. Since our problem involves a nontrivial state process, the DM who implements the $\mu$ policy may end up in a ``bad'' part of the state space, leading to an unavoidable, positive penalty. 
After period $n$, the DM cannot do any better than $\mu^\theta$.

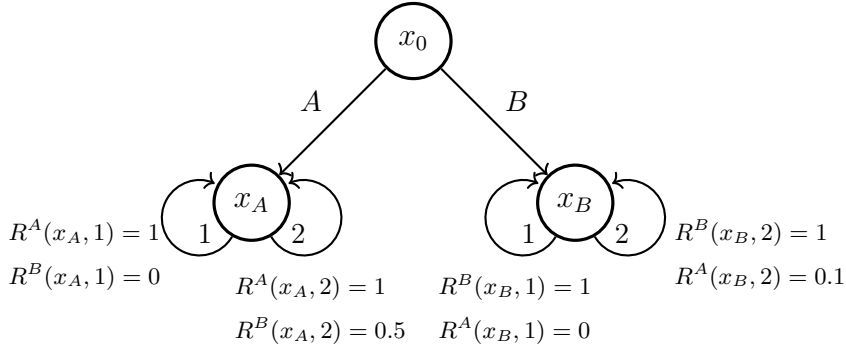
\begin{figure}[!htb]
	\begin{center}
		\begin{tikzpicture}[
		roundnode/.style={circle, draw=black, very thick, minimum size=10mm,node distance = 2cm}, 
		]
		\node[roundnode] (maintopic) {\large $x_0$};
		\node[roundnode] (lowerleftcircle) [below left=of maintopic] 
		{\large $x_A$};
		\node[roundnode] (lowerrightcircle) [below right=of maintopic] 
		{\large $x_B$};
		\draw[thick,->] (maintopic.south west) -- (lowerleftcircle.north east) node[pos=0.5,above left] {$A$};
		\draw[thick,->] (maintopic.south east) -- (lowerrightcircle.north west) node[pos=0.5,above right] {$B$};
		\draw[thick,->] (lowerleftcircle.-120) arc (-30:-30+-264:5mm) node[pos=0.3,left] {\footnotesize{
				\begin{tabular}{l}	
				$R^A(x_A,1)=1$ \\
				$R^B(x_A,1)=0$
				\end{tabular}}} 
		node[pos=0.3,above right] {1} (lowerleftcircle);
		\draw[thick,->] (lowerleftcircle.-60) arc (210:210+264:5mm) node[pos=0.3,below] {\footnotesize{
				\begin{tabular}{l}	
				$R^A(x_A,2)=1$ \\
				$R^B(x_A,2)=0.5$
				\end{tabular}}} 
		node[pos=0.3,above left] {2} (lowerleftcircle);
		\draw[thick,->] (lowerrightcircle.-60) arc (210:210+264:5mm) node[pos=0.3,right] {\footnotesize{
				\begin{tabular}{l}	
				$R^B(x_B,2)=1$ \\
				$R^A(x_B,2)=0.1$
				\end{tabular}}}
		node[pos=0.3,above left] {2} (lowerrightcircle);
		\draw[thick,->] (lowerrightcircle.-120) arc (-30:-30+-264:5mm) node[pos=0.3,below] {\footnotesize{
				\begin{tabular}{l}	
				$R^B(x_B,1)=1$ \\
				$R^A(x_B,1)=0$
				\end{tabular}}} 
		node[pos=0.3,above right] {1} (lowerrightcircle);
		\end{tikzpicture}
		\caption{Deterministic reward depending on the control, where true parameter is B.} \label{stateRegretFig}
	\end{center}
\end{figure}


\begin{example}[Absorption into an unfavorable set of states] \label{statePenaltyExample}
	Consider Figure \ref{stateRegretFig}. 
	In period $n=0$, the DM can choose either $A$ or $B$ with an immediate reward of 0. After being absorbed into one of $x_A$ or $x_B$, the DM can pick either control 1 or 2 and receives a deterministic reward, as a function of the control and true parameter. Suppose the true parameter $\theta$ is $B$, which is unknown to the DM who performs TS.
\end{example} 


If $\pi(A\mid x_0)\approx 1$, then the DM would initially sample $A$, and consequently would stay in $x_A$ forever. The $\mu$ policy may then pick control 1 (generating a reward of 0) or pick control 2 (generating a reward of 0.5). Recall that the system designer provides the transition and reward densities, i.e., the reward structures $R^A(\cdot)$ and $R^B(\cdot)$ are both known to the DM. Since the rewards are deterministic and have different values, regardless of the control picked at $n=1$, the $\mu$ policy will immediately learn that $\theta = B$. Hence, $\mu$ will always pick control 2 from $n=2$ onward, incurring a reward of 0.5. The $\theta$-optimal policy will pick parameter $B$ in period $n=0$, receiving a reward of 1 forever. Then, by (\ref{statePenalty}), the undiscounted expected total regret of being in state $x_A$, as opposed to $x_B$, increases linearly, by $1-0.5 = 0.5$ in each period. In contrast, the final term (\ref{residualRegret1}) equals 0 when $n\ge 2$. Once landing in state $x_A$, after 1 period both TS and an optimal policy will choose the same control, i.e., control 2, forever. However, (\ref{residualRegret1}) does not always converge to 0 as rapidly. In the next section, we illustrate that when the rewards are not deterministic, the $\mu$ policy learns more slowly. 

\subsection{Expected Residual Regret\label{residualSection}} 
Definition \ref{def:expected_residual_regret} formally defines component (\ref{residualRegret1}) of regret. Similar to the expected state regret, we only denote the starting period $n$ inside the parenthesis.
Consider the first term of (\ref{ResidualRegret}), 
$$
\mathbb{E}_{x_0}^{\mu,\theta}[\nu^\theta(X_n)] = \mathbb{E}_{x_0}^{\mu,\theta} \left[\sup_{\mu\in \mathcal{M}} V_{X_n}^{\mu,\theta}(0)\right] = \mathbb{E}_{x_0}^{\mu,\theta} \left[\sup_{\mu\in \mathcal{M}} \mathbb{E}_{X_n}^{\mu,\theta} \left[\sum_{t=0}^\infty\beta^{t}R_t \right]\right].
$$

The $\theta$-optimal policy starts from a random state $X_n$, which is driven by running the $\mu$ policy for $n$ periods, starting at $x_0$. 
The expectation is taken over all paths leading to all possible $X_n$. 
Hence, $\mathbb{E}_{x_0}^{\mu,\theta}[\nu^\theta(X_n)]$ is a deterministic quantity.
The second term of (\ref{ResidualRegret}) coincides with (\ref{V(n)}). 
The expectation operator is induced by $\mu$, the starting state $x_0$ and the true parameter $\theta$. Hence, similar to the first term, $V_{x_0}^{\mu,\theta}(n)$ does not forget the past; periods $0$ to $(n-1)$ impact the state wherein the process finds itself in period $n$.
Both $\mathbb{E}_{x_0}^{\mu,\theta}[\nu^{\theta}(X_n)]$ and $V_{x_0}^{\mu,\theta}(n)$ discard the rewards generated during the first $n$ periods; however, they are not independent of the past, since the random state $X_n$ is driven by the tuple ($x_0,\mu,\theta$). 
In summary, we decompose the standard regret into three components, i.e.
\begin{align} \label{decomposition}
E[\textnormal{Regret}^\theta(0,\infty)] = E[\textnormal{Regret}^\theta_n(0,n-1)] + \mathcal{S}_{x_0}^{\mu,\theta}(n) + \mathcal{R}_{x_0}^{\mu,\theta}(n).
\end{align}

Note that in (\ref{decomposition}) $\mathcal{S}_{x_0}^{\mu,\theta}(n)$ and $\mathcal{R}_{x_0}^{\mu,\theta}(n)$ tally the difference in rewards starting from $n$ into infinity, while $E[\textnormal{Regret}^\theta(0,\infty)]$ and $E[\textnormal{Regret}^\theta_n(0,n-1)]$ specify both the starting and ending periods inside the parenthesis. As previously explained, $E[\textnormal{Regret}^\theta_n(0,n-1)]$ is the accumulation of the past losses, before period $n$. This component of the expected regret is sunk, in the sense that the DM cannot change it starting from period $n$. Similarly, the expected state regret $\mathcal{S}_{x_0}^{\mu,\theta}(n)$ is the accumulation of the future losses, as a result of irrevocably being in a given state in period $n$, thus is also sunk. $E[\textnormal{Regret}^\theta_n(0,n-1)]$ and $\mathcal{S}_{x_0}^{\mu,\theta}(n)$ measure the past and future consequences, respectively, of adopting a sampling-based algorithm from period $0$ up until period $n$. From the perspective of a DM who is already in period $n$, neither can be influenced. In contrast, we regard the expected residual regret $\mathcal{R}_{x_0}^{\mu,\theta}(n)$ as ``controllable'' because, starting in period $n$, a DM can choose a policy other than $\mu$ from that period onward. Therefore, amongst the three components, only expected residual regret represents the efficacy of continuing with $\mu$ into the future.

\paragraph{\bf Illustration of Expected Residual Regret\label{residualRegretSection}} 
Now we illustrate how the expected residual regret quantifies the effectiveness of future decisions through an example. If the DM picks the best control(s) from period $n$ onward, no matter how unfavorable the state $X_n$ is, the expected residual regret starting in that period is 0. To illustrate how learning occurs in a setting with stochastic rewards, consider a single-state $\theta$-MDP example. We show how the expected residual regret is driven down to 0 as a result of $\mu$ learning over time.

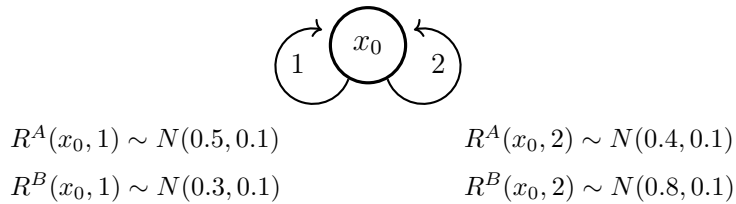
\begin{figure}[htb!]
	\begin{center}
		\begin{tikzpicture}[
		roundnode/.style={circle, draw=black, very thick, minimum size=10mm,node distance = 3cm}, 
		]
		\node[roundnode] (maintopic) {\large $x_0$};
		\draw[thick,->] (maintopic.-120) arc (-20:-20+-264:5mm) node[pos=0.3,below left] {\small{
				\begin{tabular}{l}	
				$R^A(x_0,1) \sim N(0.5,0.1)$ \\
				$R^B(x_0,1) \sim N(0.3,0.1)$
				\end{tabular}}} 
		node[pos=0.5,above right] {1} (maintopic);
		\draw[thick,->] (maintopic.-60) arc (200:200+264:5mm) node[pos=0.3,below right] {\small{
				\begin{tabular}{l}	
				$R^A(x_0,2) \sim N(0.4,0.1)$ \\
				$R^B(x_0,2) \sim N(0.8,0.1)$
				\end{tabular}}}
		node[pos=0.5,above left] {2} (maintopic);
		\end{tikzpicture}
		\caption{Stochastic rewards depending on the control, where the true parameter is B.} \label{residualRegretFig}
	\end{center}
\end{figure}

\begin{example}[Expected residual regret converges to 0] \label{residualRegretExample}
	Consider the single-state $\theta$-MDP in Figure \ref{residualRegretFig}. The underlying reward structure is illustrated on the arcs, which represent the controls. 
	The rewards are stochastic. 
	Based on the true parameter (which we assume is $B$) and the control picked at every step, the reward is generated from a normal distribution with known mean and variance. Not knowing the current state, the DM draws a sample from the posterior distribution in each period and obtains either $\Theta_t = A$ or $\Theta_t = B$. The initial prior belief on $A$ is 0.5. In each step of the process, if the sample drawn is $A$, then the DM picks control 1. This is because control 1 has an expected reward of 0.5 when sampling $A$, while control 2 has an expected reward of 0.4. On the other hand, if the DM samples $B$, then they pick control 2 since the corresponding expected reward is higher.
\end{example}

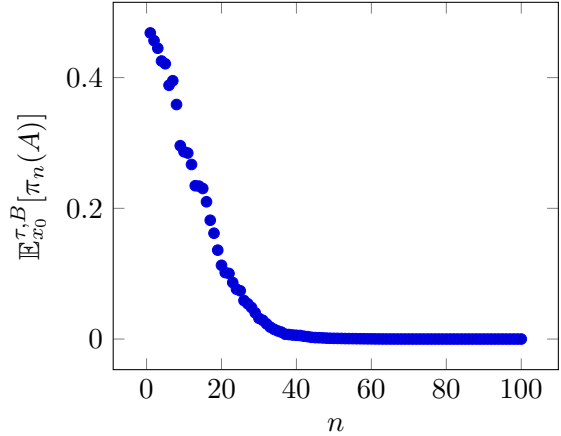
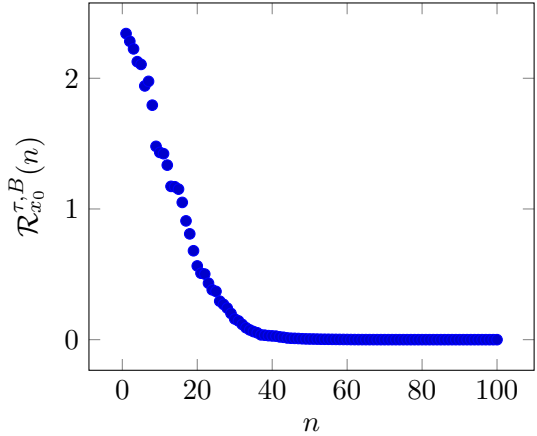
\begin{figure}[!htb]
	\centering
	\begin{subfigure}[t]{.48\linewidth}
		\centering
		\begin{tikzpicture}
		\begin{axis}[
		width=\linewidth,
		xlabel={$n$},
		ylabel={$\mathbb{E}_{x_0}^{\mu,B}[\pi_n(A)]$}
		]
		
		\addplot
		coordinates {
			(1,0.468382)
			
			(2,0.456574)
			
			(3,0.444958)
			
			(4,0.4254)
			
			(5,0.42118)
			
			(6,0.388291)
			
			(7,0.395253)
			
			(8,0.358826)
			
			(9,0.295819)
			
			(10,0.286599)
			
			(11,0.284577)
			
			(12,0.267082)
			
			(13,0.234579)
			
			(14,0.233989)
			
			(15,0.23037)
			
			(16,0.210036)
			
			(17,0.181772)
			
			(18,0.161903)
			
			(19,0.136041)
			
			(20,0.112836)
			
			(21,0.101721)
			
			(22,0.100379)
			
			(23,0.0864483)
			
			(24,0.0763013)
			
			(25,0.0738714)
			
			(26,0.0589434)
			
			(27,0.0540751)
			
			(28,0.0483213)
			
			(29,0.0401627)
			
			(30,0.0317393)
			
			(31,0.0287436)
			
			(32,0.023348)
			
			(33,0.0183409)
			
			(34,0.0149725)
			
			(35,0.0124413)
			
			(36,0.0105448)
			
			(37,0.0073939)
			
			(38,0.0068876)
			
			(39,0.00616146)
			
			(40,0.0057457)
			
			(41,0.00542329)
			
			(42,0.00428366)
			
			(43,0.00366576)
			
			(44,0.0026767)
			
			(45,0.0022252)
			
			(46,0.0020275)
			
			(47,0.00176491)
			
			(48,0.0014124)
			
			(49,0.00118843)
			
			(50,0.00106896)
			
			(51,0.000999799)
			
			(52,0.000873761)
			
			(53,0.000704149)
			
			(54,0.000648864)
			
			(55,0.000605109)
			
			(56,0.000509376)
			
			(57,0.000454235)
			
			(58,0.000382716)
			
			(59,0.000324899)
			
			(60,0.000278666)
			
			(61,0.000191307)
			
			(62,0.000137334)
			
			(63,0.000104376)
			
			(64,0.0000823055)
			
			(65,0.0000708703)
			
			(66,0.0000615567)
			
			(67,0.0000519644)
			
			(68,0.0000456343)
			
			(69,0.0000373332)
			
			(70,0.0000328058)
			
			(71,0.0000275503)
			
			(72,0.0000259234)
			
			(73,0.0000237644)
			
			(74,0.0000219847)
			
			(75,0.0000199145)
			
			(76,0.0000177036)
			
			(77,0.0000142013)
			
			(78,0.0000138854)
			
			(79,0.0000110693)
			
			(80,0.00000915722)
			
			(81,0.00000787858)
			
			(82,0.00000682556)
			
			(83,0.00000534262)
			
			(84,0.00000457551)
			
			(85,0.00000365197)
			
			(86,0.00000340814)
			
			(87,0.00000309289)
			
			(88,0.00000254502)
			
			(89,0.0000020838)
			
			(90,0.00000184792)
			
			(91,0.00000161055)
			
			(92,0.00000136338)
			
			(93,0.00000115182)
			
			(94,0.000000981798)
			
			(95,0.000000882605)
			
			(96,0.000000811632)
			
			(97,0.000000657511)
			
			(98,0.00000060552)
			
			(99,0.000000551054)
			
			(100,0.000000515626)
		};
		\end{axis}
		\end{tikzpicture} 
		\caption{Evolution of expected posterior sampling error}
		\label{fig:Ng1}
	\end{subfigure}
	\hfill
	\begin{subfigure}[t]{.48\linewidth}
		\centering
		\begin{tikzpicture}
		\begin{axis}[
		width=\linewidth,
		xlabel={$n$},
		ylabel={$\mathcal{R}_{x_0}^{\mu,B}(n)$}]
		\addplot coordinates {
			(1,2.34191)
			
			(2,2.28287)
			
			(3,2.22479)
			
			(4,2.127)
			
			(5,2.1059)
			
			(6,1.94145)
			
			(7,1.97626)
			
			(8,1.79413)
			
			(9,1.47909)
			
			(10,1.43299)
			
			(11,1.42289)
			
			(12,1.33541)
			
			(13,1.1729)
			
			(14,1.16994)
			
			(15,1.15185)
			
			(16,1.05018)
			
			(17,0.908862)
			
			(18,0.809516)
			
			(19,0.680205)
			
			(20,0.564178)
			
			(21,0.508606)
			
			(22,0.501894)
			
			(23,0.432241)
			
			(24,0.381506)
			
			(25,0.369357)
			
			(26,0.294717)
			
			(27,0.270376)
			
			(28,0.241606)
			
			(29,0.200813)
			
			(30,0.158697)
			
			(31,0.143718)
			
			(32,0.11674)
			
			(33,0.0917045)
			
			(34,0.0748623)
			
			(35,0.0622067)
			
			(36,0.0527241)
			
			(37,0.0369695)
			
			(38,0.034438)
			
			(39,0.0308073)
			
			(40,0.0287285)
			
			(41,0.0271165)
			
			(42,0.0214183)
			
			(43,0.0183288)
			
			(44,0.0133835)
			
			(45,0.011126)
			
			(46,0.0101375)
			
			(47,0.00882453)
			
			(48,0.00706199)
			
			(49,0.00594217)
			
			(50,0.00534479)
			
			(51,0.00499899)
			
			(52,0.0043688)
			
			(53,0.00352074)
			
			(54,0.00324432)
			
			(55,0.00302554)
			
			(56,0.00254688)
			
			(57,0.00227117)
			
			(58,0.00191358)
			
			(59,0.00162449)
			
			(60,0.00139333)
			
			(61,0.000956533)
			
			(62,0.00068667)
			
			(63,0.00052188)
			
			(64,0.000411527)
			
			(65,0.000354351)
			
			(66,0.000307783)
			
			(67,0.000259822)
			
			(68,0.000228171)
			
			(69,0.000186666)
			
			(70,0.000164029)
			
			(71,0.000137752)
			
			(72,0.000129617)
			
			(73,0.000118822)
			
			(74,0.000109924)
			
			(75,0.0000995724)
			
			(76,0.0000885179)
			
			(77,0.0000710067)
			
			(78,0.0000694271)
			
			(79,0.0000553464)
			
			(80,0.0000457861)
			
			(81,0.0000393929)
			
			(82,0.0000341278)
			
			(83,0.0000267131)
			
			(84,0.0000228776)
			
			(85,0.0000182599)
			
			(86,0.0000170407)
			
			(87,0.0000154645)
			
			(88,0.0000127251)
			
			(89,0.000010419)
			
			(90,0.00000923961)
			
			(91,0.00000805277)
			
			(92,0.00000681688)
			
			(93,0.00000575912)
			
			(94,0.00000490899)
			
			(95,0.00000441302)
			
			(96,0.00000405816)
			
			(97,0.00000328756)
			
			(98,0.0000030276)
			
			(99,0.00000275527)
			
			(100,0.00000257813)
		};
		\end{axis}
		\end{tikzpicture}
		\caption{Evolution of expected residual regret}
		\label{fig:Ng2}
	\end{subfigure}
	\caption{Evolution of expected posterior and expected residual regret in Example \ref{residualRegretExample} when first sample is wrong}
	\label{label1}
\end{figure}

Figure \ref{fig:Ng1} illustrates the evolution of the expected belief on the ``wrong'' parameter $\mathbb{E}_{x_0}^{\mu,B}[\pi_n(A)]$, i.e., the expected posterior sampling error, averaged over 100 runs of the TS policy. We observe that the DM learns the true parameter $B$ not immediately, but after approximately 40 iterations. Beyond that point, the DM always picks control 2, in order to maximize the infinite-horizon expected total reward. Figure \ref{fig:Ng2} illustrates the decline in the expected residual regret for the same example.

Consider a hybrid policy that switches from $\mu$ to $\mu^B$ after running $\mu$ for $n$ periods. 
Since the true parameter is $B$, the reward distributions follow $N(0.3,0.1)$ and $N(0.8,0.1)$ when control 1 or 2 is selected, respectively, i.e., the realizations of $R^B(x_0,1)$ and $R^B(x_0,2)$ are the rewards associated with control 1 and 2, respectively. Therefore, the expected reward of control 2 is greater than the expected reward of control 1 and the $\mu^B$ policy always picks control 2, i.e., 
$$
\int r f^B(x_0,2) \, dr = 0.8 > \int r f^B(x_0,1) \, dr = 0.3.
$$

From period $n$ onward, a hybrid policy gains total expected reward of 
\begin{align*}
\mathbb{E}_{x_0}^{\mu,B}[\nu^{B}(X_n)] &= \mathbb{E}_{x_0}^{\mu,B} \left[\mathbb{E}_{X_n}^{\mu^B, B} \left[\sum_{t=0}^\infty \beta^t R_t\right]\right] \\
& = \sum_{t=0}^\infty\beta^t0.8 = \frac{0.8}{1-\beta} = \nu^B(x_0).
\end{align*}

We emphasize that $n$ corresponds to a fixed time period, while $t$ is used when  summing over the rewards. Consider the $\mu_1$ policy that always picks control 1. Had the DM implemented such a policy, the expected residual regret would have been 
\begin{align*}
\mathcal{R}_{x_0}^{\mu_1,B}(n) & =\mathbb{E}_{x_0}^{\mu_1,B} [\nu^B (X_n)] - V_{x_0}^{\mu_1,B}(n) \\
& = \frac{0.8}{1-\beta} - \mathbb{E}_{x_0}^{\mu_1,B} \left[\sum_{t=n}^\infty \beta^{t-n} R_t\right] \\
& = \sum_{t=0}^\infty \beta^t 0.8 - \sum_{t=n}^\infty \beta^{t-n} 0.3 = \sum_{t=0}^\infty \beta^t 0.5 = \frac{0.5}{1-\beta}.
\end{align*}

Thus, $\frac{0.5}{1-\beta}$ is an upper bound on the expected residual regret for all $n$. After some number of periods, the rate at which the policy $\mu$ picks control 1 becomes negligible. Once the policy no longer picks control 1, the DM has figured out the true parameter is $B$. If we call this period $\tilde{n}$, the expected residual regret becomes
$$
\mathcal{R}_{x_0}^{\mu,B}(n) \coloneqq \mathbb{E}_{x_0}^{\mu,B} [\nu^B (X_n)] - V_{x_0}^{\mu,B}(n) = \sum_{t=0}^\infty \beta^t 0.8 - \sum_{t=n}^\infty \beta^{t-n}  0.8 = 0, \quad \forall n\ge \tilde{n}.
$$

Starting from $n=0$ and using Example \ref{residualRegretExample} assumptions, the expected residual regret can be computed as 
$$
\mathcal{R}_{x_0}^{\mu,B}(n)= \sum_{t=0}^\infty \beta^t 0.8 - \left(\mathbb{E}_{x_0}^{\mu,B}[\pi_n(A)]\frac{0.3}{1-\beta} + \mathbb{E}_{x_0}^{\mu,B}[\pi_n(B)]\frac{0.8}{1-\beta}\right).
$$

Figure \ref{fig:Ng2} shows the evolution of $\mathcal{R}_{x_0}^{\mu,B}$ for $\beta=0.9$. When a different $\beta\in [0,1)$ is chosen, the trajectory remains the same, but the expected residual regret values are different for small $n$.

\subsection{Probabilistic Residual Regret\label{probabilisticResidualRegret}} 
Definition \ref{def:residual_regret} defines the residual regret as the ``probabilistic'' version of the expected residual regret, $\mathcal{R}_{x_0}^{\tau,\theta}(n)$. Recall the interpretation of the first term of (\ref{resRegDiff3}); the stochastic process is driven by a ``hybrid'' policy that follows the $\mu$ policy for the first $n$ periods, and then adopts the optimal policy $\mu^\theta$ in period $n$ after the oracle reveals $\theta$. It is a random quantity due to the random starting state $X_n$ in period $n$, induced by the learning algorithm. The second term is the random expected infinite sum obtained by implementing the sampling-based policy $\mu$, starting from random state $X_n$, and omitting the rewards from period $0$ to $n-1$.

\color{black}

\section{Analysis of Assumption \ref{relativeEntropy}} \label{AppendixAssumptionVerif}

Recall Assumption \ref{relativeEntropy}. Since Example \ref{residualRegretExample} has a finite state space, (\ref{entropy_bounded_away}) boils down to: For any $x\in\mathcal{X}$, $u\in\mathcal{U}$, and any two distinct parameter value, $\gamma\ne\theta\in\mathcal{P}$, there exists a positive constant $\epsilon(x,u,\theta,\gamma) > 0$ such that
$$
\mathbb{K}(\nu_\theta^{x,u}\mid \nu_\gamma^{x,u})\ge \epsilon(x,u,\theta,\gamma).
$$

The above condition holds if and only if (by definition)
$$
E_{f^\theta q^\theta}\left[\log\left(\frac{d\nu_\theta^{x,u}}{d\nu_\gamma^{x,u}}\right)\right] = E_{f^\theta q^\theta}\left[\log\left(\frac{f^\theta(\cdot\mid x,u)q^\theta(\cdot\mid x,u)}{f^\gamma(\cdot\mid x,u)q^\gamma(\cdot\mid x,u)}\right)\right] \ge \epsilon(x,u,\theta,\gamma),
$$
where $E_{f^\theta q^\theta}[\cdot\mid x,u]$ is defined in Section \ref{ProblemFormulation}. For illustration purposes, we make the same assumption we had in Example \ref{residualRegretExample}; the $\theta$-MDP has only one state. This implies the transition kernel, $q^\theta(\cdot\mid x,u)$, is deterministic and we simplify the relative entropy expression. Due to the simpler version of the relative entropy, we utilize the expectation operator $E_{f^\theta}[\cdot\mid x,u]$, also defined in Section \ref{ProblemFormulation}. It suffices to find a constant $\epsilon(x,u,\theta,\gamma)>0$, for any $x_A$ (in this case $\mathcal{X} = \{x_0\}$ where $x_0=x_A$), any $u\in\mathcal{U}$ (control 1 or 2), $\theta = B$ and $\gamma = A$, such that 
$$
E_{f^\theta}\left[\log\left(\frac{f^\theta(\cdot\mid x,u)}{f^\gamma(\cdot\mid x,u)}\right)\right] \ge \epsilon(x,u,\theta,\gamma).
$$

We have,
\begin{align*}
E_{f^\theta}\left[\log\left(\frac{\frac{1}{\sqrt{2\pi 0.1}}e^{-\frac{(r-\mu^\theta)^2}{2(0.1)}}\Big\vert x,u}{\frac{1}{\sqrt{2\pi 0.1}}e^{-\frac{(r-\mu^\gamma)^2}{2(0.1)}}\Big\vert x,u}\right)\right] &= E_{f^\theta}\left[\log\left(\frac{e^{-(r-\mu^\theta)^2}\big\vert x,u}{e^{-(r-\mu^\gamma)^2}\big\vert x,u}\right)\right] \\
&= E_{f^\theta}\left[\log\left(e^{-(r-\mu^\theta)^2 + (r-\mu^\gamma)^2}\big\vert x,u\right)\right],
\end{align*}
which can be simplified as
$$
E_{f^\theta}[-(r-\mu^\theta)^2 + (r-\mu^\gamma)^2\mid x,u].
$$

\noindent \textbf{Case 1:} When the state is $x_A$ and control 2 is picked by the TS policy, $r\sim N(0.8,0.1)$, and
\begin{align*}
&E_{f^\theta}[-(r-0.8)^2 + (r-0.4)^2 \mid x_A,2] \\
&= - Var[r\mid x_A,2] + E_{f^\theta} [(r - 0.4)^2 \mid x_A,2] \\
&= - 0.1 + E_{f^\theta} [r^2 - 0.8r + 0.16 \mid x_A,2] \\
&= - 0.1 + E_{f^\theta} [r^2 \mid x_A,2] - 0.8E_{f^\theta}[r\mid x_A,2] + 0.16 \\
&= - 0.1 + Var[r\mid x_A,2] + (E_{f^\theta} [r\mid x_A,2])^2 - 0.8E_{f^\theta}[r\mid x_A,2] + 0.16 \\
&= -0.1 + 0.1 + 0.64 - 0.64 + 0.16 = 0.16.
\end{align*}
\textbf{Case 2:} When the state is $x_A$ and control 1 is picked by the TS policy, $r\sim N(0.3,0.1)$, and
\begin{align*}
&E_{f^\theta}[-(r-0.3)^2 + (r-0.5)^2 \mid x_A,1] \\
&= - Var[r\mid x_A,1] + E_{f^\theta} [(r - 0.5)^2 \mid x_A,1] \\
&= - 0.1 + E_{f^\theta} [r^2 - r + 0.25 \mid x_A,1] \\
&= - 0.1 + E_{f^\theta} [r^2 \mid x_A,1] - E_{f^\theta}[r\mid x_A,1] + 0.25 \\
&= - 0.1 + Var[r\mid x_A,1] + (E_{f^\theta} [r\mid x_A,1])^2 - E_{f^\theta}[r\mid x_A,1] + 0.25 \\
&= -0.1 + 0.1 + 0.09 - 0.3 + 0.25 = 0.04.
\end{align*}

As long as $0 < \epsilon(x,u,\theta,\gamma) \le 0.04$, Assumption \ref{relativeEntropy} holds for Example \ref{residualRegretExample}.


\bibliographystyle{informs2014}

\bibliography{TS}

\end{document}